*Heaven's Light is Our Guide*

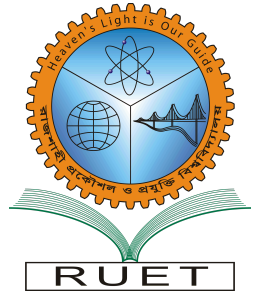

**DEPARTMENT OF COMPUTER SCIENCE & ENGINEERING**
**RAJSHAHI UNIVERSITY OF ENGINEERING & TECHNOLOGY, BANGLADESH**

**SMFD-UNet: Semantic Face Mask Is The Only Thing You Need To Deblur Faces**

**Author**
Abduz Zami
Roll No. 1903158
Department of Computer Science & Engineering
Rajshahi University of Engineering & Technology

**Supervised by**
Md. Nasif Osman Khansur
Lecturer
Department of Computer Science & Engineering
Rajshahi University of Engineering & Technology

# ACKNOWLEDGEMENT

With profound gratitude and unwavering belief in divine grace, I acknowledge the Almighty's guidance throughout my thesis journey. Without His benevolence, this task would have been insurmountable. May His wisdom continue to lead us.

In this endeavor, I am immensely indebted to the esteemed **Md. Nasif Osman Khansur**, Lec-turer, Department of Computer Science & Engineering, Rajshahi University of Engineering & Technology, Rajshahi. His unwavering support and invaluable assistance throughout this study have been instrumental in its successful completion. At every step, his wise counsel and en-couragement motivated me to strive for excellence and evolve into an independent researcher. His compassion and generosity served as a source of solace during times of mental strain. I consider myself incredibly fortunate to have him as our supervisor, and his inspiration has been the driving force behind this thesis.

My sincere gratitude extends to the entire faculty, whose precious time, expertise, and efforts have created an environment conducive to academic research. Their dedication to nurturing our intellect and fostering our growth is deeply appreciated.

Finally, I extend my heartfelt appreciation to my beloved parents, whose unwavering love and unwavering support have been the pillars of strength throughout my academic journey.

July 2, 2025                                                              Abduz Zami
RUET, Rajshahi

*Heaven's Light is Our Guide*

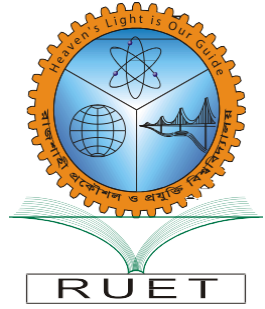

# DEPARTMENT OF COMPUTER SCIENCE & ENGINEERING
# RAJSHAHI UNIVERSITY OF ENGINEERING & TECHNOLOGY, BANGLADESH

# *CERTIFICATE*

*This is to certify that this thesis report entitled **"SMFD-UNet: Semantic Face Mask Is The Only Thing You Need To Deblur Faces"** submitted by **Abduz Zami, Roll:1903158** in partial fulfillment of the requirement for the award of the degree of Bachelor of Science in Department of Computer Science & Engineering of Rajshahi University of Engineering & Technology is a record of the candidates own work carried out by them under my supervision. This thesis has not been submitted for the award of any other degree.*

Supervisor                                              External Examiner

———————————————                    ———————————————

**Md. Nasif Osman Khansur**                             **Md. Rokanujjaman**
Lecturer                                                Professor
Department of Computer Science &                        Department of Computer Science &
Engineering                                             Engineering
Rajshahi University of Engineering &                    Rajshahi University
Technology

Rajshahi-6204                                           Rajshahi-6205

# ABSTRACT

For applications including facial identification, forensic analysis, photographic improvement, and medical imaging diagnostics, facial image deblurring is an essential chore in computer vi-sion allowing the restoration of high-quality images from blurry inputs. Often based on general picture priors, traditional deblurring techniques find it difficult to capture the particular struc-tural and identity-specific features of human faces. We present SMFD-UNet (Semantic Mask Fusion Deblurring UNet), a new lightweight framework using semantic face masks to drive the deblurring process, therefore removing the need for high-quality reference photos in order to solve these difficulties. First, our dual-step method uses a UNet-based semantic mask generator to directly extract detailed facial component masks (e.g., eyes, nose, mouth) straight from blurry photos. Sharp, high-fidelity facial images are subsequently produced by integrating these masks with the blurry input using a multi-stage feature fusion technique within a computationally ef-ficient UNet framework. We created a randomized blurring pipeline that roughly replicates real-world situations by simulating around 1.74 trillion deterioration scenarios, hence guar-anteeing resilience. Examined on the CelebA dataset, SMFD-UNet shows better performance than state-of-the-art models, attaining higher Peak Signal-to-Noise Ratio (PSNR) and Structural Similarity Index Measure (SSIM) while preserving satisfactory naturalness measures, including NIQE, LPIPS, and FID. Powered by Residual Dense Convolution Blocks (RDC), a multi-stage feature fusion strategy, efficient and effective upsampling techniques, attention techniques like CBAM, post-processing techniques, and the lightweight design guarantees scalability and effi-ciency, enabling SMFD-UNet to be a flexible solution for developing facial image restoration research and useful applications.

**Keywords:** Image Deblurring, Image Restoration, Multi-Stage Fusion, UNet

# CONTENTS

# LIST OF TABLES

# LIST OF FIGURES

# LIST OF ABBREVIATIONS

| Abbreviation | Description |
| --- | --- |
| SMFD | Semantic Mask Face Deblurring |
| UNet | U-shaped Network |
| CNN | Convolutional Neural Network |
| CBAM | Convolutional Block Attention Module |
| RDC | Residual Dense Convolution Block |
| PSNR | Peak Signal-to-Noise Ratio |
| SSIM | Structural Similarity Index Measure |
| GAN | Generative Adversarial Network |
| ReLU | Rectified Linear Unit |
| LPIPS | Learned Perceptual Image Patch Similarity |
| MSE | Mean Square Error |
| NIQE | Natural Image Quality Evaluator |
| FID | Fréchet Inception Distance |

# Chapter 1

# Introduction

## 1.1 Introduction

Deblurring is the process of removing blurriness from an image. When it comes to the face, it is referred to as facial image deblurring. Blurriness in images can result from camera motion, defocus, or unfavorable ambient conditions, among other things, and such degradations remove important identity-specific characteristics, hence affecting downstream task performance [1]. The most common forms of blur are motion and gaussian. Figure 1.2 shows how these two forms might create blurry facial images. Facial images shown in this figure and all other facial images throughout this thesis book are taken from the CelebA-Mask-HQ dataset [2] unless explicitly stated.

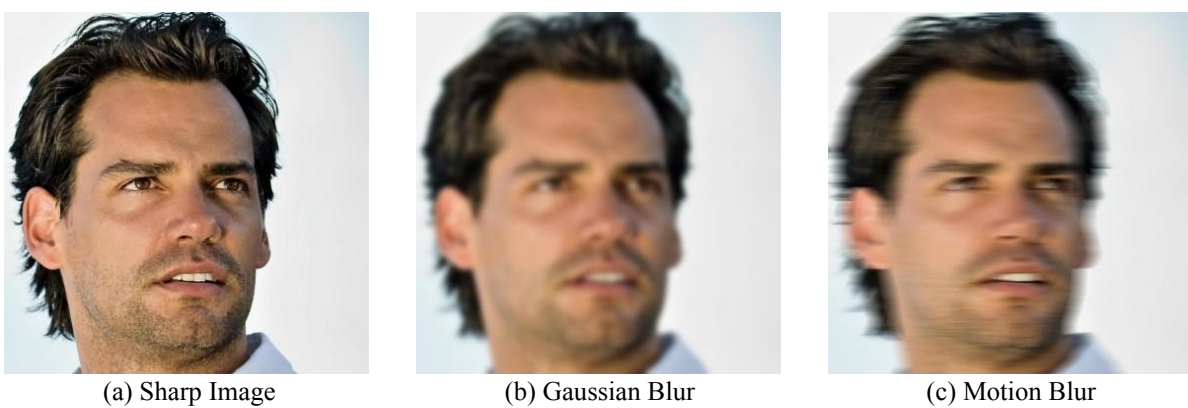

(a) Sharp Image　　　　　(b) Gaussian Blur　　　　　(c) Motion Blur

Figure 1.1: Comparison Of A Sharp Image and Its Blurred Versions

Semantic face masks are applied in this work to preserve the geometric forms after deblur-ring. A semantic face mask is the depiction of areas and locations of several facial components such as hair, nose, eye, eyebrows, ear, or any other wearings. Figure 1.2 provides a semantic face mask example.

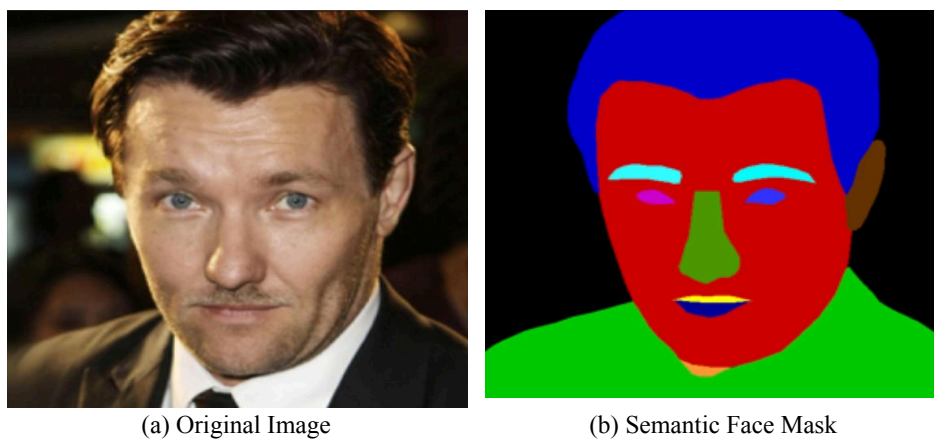

(a) Original Image      (b) Semantic Face Mask

Figure 1.2: Original Image and Semantic Face Mask of the Image

The subsequent section of this chapter begins with a summary of the work that was done for the thesis. This thesis discusses the rationale behind its topic and outlines the research aims. A comprehensive overview of the problem is provided. An analysis of a user's needs is performed. The next step is to provide an explanation of the study's aims, questions, hypotheses, and the most important contribution that this thesis work makes. It is also mentioned how the job done for this thesis will affect the real world. Sustainable environmental practices, ethical considerations, and the timetable for the completion of this thesis are also mentioned. At the end, it clarifies the book's structure and its conclusion.

## 1.2 Motivation

In many fields, including facial recognition, forensic investigations, photography enhancement, and medical imaging [3, 4, 5, 6], facial image deblurring techniques are indispensable. Tradi-tional single-image deblurring methods, typically rooted in the maximum a posteriori (MAP) framework, rely on generic natural image priors—such as sparsity in gradient domains or edge distributions—to constrain the solution space [7].

While deep learning has advanced general image deblurring, these techniques often under-perform on facial images due to their unique structural characteristics [8]. To address this, re-searchers have proposed specialized approaches for facial deblurring. For instance, generative adversarial networks (GANs) and diffusion models [9, 10] are proposed by studies with adver-sarial and perceptual losses. However, these often oversharpen images, altering identity char-acteristics and yielding unrealistic results. Transformer-based methods have shown promise but remain computationally intensive compared to traditional convolutional neural networks [11, 12, 13].

Alternative strategies include using reference images of the same identity [10]. Finding a reference image is not always feasible. Some approaches involve constructing dictionaries of facial components from diverse identities to assist deblurring [14]. But this process is more complex and requires a massive collection of reference images, facial component data, and mapping.

Single semantic-guided methods, for instance, leverage facial features such as edges, hair-lines, or component masks (e.g., eyes, ears, eyebrows) and even 3D landmarks to guide restora-tion [15, 16, 17] is a great way to preserve identity characteristics. But existing works mostly rely on reference images for semantics. Few researchers, such as [15] suggested a method for producing semantic masks from blurry images, but it has limitations, such as failing when the input image is not well aligned and for extremely excessive motion blur.

This thesis's main driving force is addressing important research gaps in image deblurring, including the computational complexity of current methods, the problem of synthetic image generation, the need for reference images, and the indispensable function of deblurring across many uses. Furthermore, this work aims to address the current research gap by utilizing seman-tic face masks to maintain structural similarity without relying on reference images, thereby preserving structural integrity. This will help to keep structural similarity intact.

## 1.3 Research Challenges

There are a few challenges to overcome in solving issues with the existing literature. The pri-mary challenges addressed in this research include:

- Restoring highly blurry facial images without introducing identity distortions.

- Generating semantic face masks directly from blurry inputs, so removing dependence on clean reference images.

- Balancing model complexity with computational efficiency for real-world deployment

- Simulating realistic blur variations to improve generalization across real-world blurriness.

## 1.4 Research Goals

The overarching goal of this research is to develop a facial image deblurring framework that:

- Preserves the identity of the subject even under severe blur conditions.
- Functions without reliance on external reference images.
- Operates efficiently within computational constraints.

## 1.5 Problem Statements

There are many problems with the existing literature. The main problem is that general deblur-ring does not work well for facial images, as facial image deblurring must keep the structural similarity intact. So, there is a huge need for a dedicated facial image deblurring technique. Though, many studies has proposed dedicated facial image deblurring approaches but they also have some limitations. This approach is primarily transformer-based and GAN-based. Some of the studies utilized reference images and a few more techniques.

Here is a list of a few limitations of the existing methodologies:

- General Deblurring methods can't restore geometric structures properly.
- Facial Image dedicated studies have been performed, but they have the following issues
    - Transformer-based approaches require a huge collection of data for training
    - GAN/Difusion-based approaches generate synthetic images
    - Dictionary/Uncertainty guided approaches require a collection of reference images or semantics, and are also not good for keeping structural similarity
    - Semantic Prior-based approaches are good at keeping the structural similarity intact, but existing works rely on exemplars

Structural similarity is crucial for facial images, and a semantic face mask is an effective way to preserve this while deblurring. However, existing literature relies on a reference image of the same identity, and those that don't rely on it suffer from being poorly aligned and excessively blurry. Therefore, the primary focus of this study is to address the need for a reference image by generating a semantic mask from a blurry image, thereby avoiding the

issues of excess blur

and poorly aligned images. Problems such as the need for a vast collection of data, synthetic image generation, and the requirement for exemplars should also be addressed.

## 1.6 User Requirement Analysis

The user requirements for a lightweight, semantic-guided, reference-free deblurring model in-tended to improve computer vision applications are described in this section. Globally speaking, the model tackles the requirement for effective and superior image restoration, especially in sit-uations where image clarity is crucial. In the industrial setting, it meets the need for quick, precise, and identity-preserving deblurring solutions in a variety of fields, such as surveillance, face analytics, and access control. To give a thorough grasp of the requirements from both a technical and application-specific standpoint, the analysis is broken up into subsections.

### 1.6.1 International Context and Technical Needs

The need for reliable image deblurring solutions has increased as a result of various industries' growing reliance on computer vision technologies. Conventional deblurring techniques fre-quently call for computationally demanding procedures or reference images, which make them unsuitable for real-time applications. In order to overcome these obstacles, the suggested model should be:

- **Lightweight**: Designed to have minimal computational overhead, it can be deployed on devices with limited resources, like mobile platforms and edge devices.

- **Reference-Free**: Removes the need for pristine reference photos, allowing it to be used in dynamic settings without them.

- **Identity Preserving**: The model should preserve the identical similarity, that is, not change identical characteristics or facial expressions.

The need for scalable and adaptable deblurring solutions that work well in a range of hard-ware setups and environmental circumstances is met by these technical characteristics. For ex-ample, a lightweight model guarantees wider accessibility and usability in areas with restricted access to high-performance computing.

## 1.6.2 User Needs and Industrial Applications

The design of the model is motivated by particular industrial needs, where image deblurring is essential to operational dependability and efficiency. Important areas of application include:

### 1.6.2.1 Face Analytics

Because blurry images can lead to misidentification or unreliable results—which are unaccept-able in applications like customer profiling or personalized marketing—clear images are es-sential for accurate facial recognition, emotion detection, and demographic analysis in face analytics. Models must preserve individual-specific features like facial contours and expres-sions through semantic guidance, restore precise facial features with high accuracy, and allow for quick, real-time processing for applications like social media platforms and retail video analysis in order to satisfy user requirements.

### 1.6.2.2 Access Control

Computer vision is used by access control systems in smart buildings or secure facilities to authenticate people using biometric scanning or facial recognition, but security may be com-promised by blurry images from motion or low light. Models must provide low-latency instanta-neous deblurring for seamless authentication experiences, demonstrate robustness by operating consistently under a variety of conditions, such as fast motion or poor lighting, and guaran-tee security compliance by preserving the integrity of biometric data without losing important identity markers in order to satisfy user requirements.

### 1.6.2.3 Observation

Clear video is necessary for surveillance systems in public areas, transit hubs, and private prop-erties to keep an eye on activity and guarantee safety; however, blurry images from camera motion or environmental conditions can make it difficult to identify people or objects. Models must provide scalability to deploy across large camera networks with varying hardware capa-bilities, deal with privacy concerns by using semantic-guided deblurring to restore important details like faces or license plates while respecting privacy constraints in non-critical areas, and achieve clarity in adverse conditions by handling complex blur patterns caused by weather, low resolution, or camera shake.

## 1.7 Research Objectives

Based on existing studies and limitations, objectives below were set for this study:

- To utilize semantic masks to preserve identity characteristics intact
- To generate a semantic mask from the blurry image, thus altering the need for exemplars
- To prevent synthetic image generation
- To handle excess blur
- Deblur faces with diverse alignments
- To keep the architecture lightweight

## 1.8 Research Questions

The following research questions guide this study, focusing on advancing image deblurring techniques through semantic understanding, identity preservation, and architectural efficiency:

- Can semantic information derived from blurry images improve deblurring performance?
- Is it possible to preserve identity without using external reference images?
- How effective is a lightweight architecture compared to larger state-of-the-art models?

## 1.9 Research Hypotheses

Before starting this research few hypotheses were proposed:

- **H1:** Semantic masks generated from blurry images can effectively guide identity-preserving deblurring.
- **H2:** Artificial uniform blurriness can mimic real-world nonuniform blurriness to a satis-factory level.
- **H3:** Reference-free semantic deblurring performs competitively with reference-based methods.

- **H4:** Lightweight CNN architectures can achieve high deblurring quality with lower com-putational costs.

## 1.10 Research Contribution

Before beginning this research, various ideas were found. But at the end the following are the important contributions of this thesis study, which advance the field of image deblurring through unique methodology, robust procedures, and efficient frameworks:

- Proposing a novel deblurring method outperforming state-of-the-art for severe blur.

- Proposing a blurring technique with 1.74 trillion variations for real-world robustness.

- Altering the need for exemplars by proposing a Semantic mask generator from blurry images.

- Proposing a lightweight framework for easy future refinements.

- Proposing a novel approach for multi-stage fusion for merging blurry images and seman-tic mask features.

## 1.11 Impact of this Research

The approach to deblurring heavily distorted facial images without requiring a reference image leverages a lightweight and efficient model. This innovation eliminates the need for exter-nal dependencies, enhancing the model's robustness and adaptability across diverse real-world scenarios, including varying lighting conditions, image qualities, and blur types. The model's minimal computational footprint enables seamless integration into resource-constrained envi-ronments, such as mobile devices, embedded systems, or edge computing platforms, making it highly versatile for practical applications. By delivering high-quality deblurring with reduced complexity, this research paves the way for accessible, on-the-go facial image restoration, with potential impacts in fields like photography, surveillance, biometric authentication, and social media, where clear facial images are critical. Furthermore, the lightweight design ensures faster processing times and lower energy consumption, contributing to sustainable and scalable solu-tions in image processing technology.

## 1.12 Environmental Sustainability

The model's lightweight architecture significantly reduces computational power consumption compared to resource-intensive transformer-based or GAN-based solutions, which often re-quire substantial hardware resources and energy. By optimizing both training and inference processes, the model achieves high efficiency, enabling faster processing with minimal energy expenditure. This efficiency not only lowers operational costs but also aligns with sustain-able AI development goals by reducing the carbon footprint associated with intensive com-putations. The streamlined design facilitates deployment on low-power devices, such as mo-bile phones, IoT systems, or edge devices, broadening accessibility to high-quality deblurring technology. Additionally, the reduced computational demands enable real-time applications in energy-sensitive environments, supporting eco-friendly advancements in fields like mobile pho-tography, real-time surveillance, and biometric systems, while maintaining robust performance across diverse scenarios.

## 1.13 Ethics

The face image deblurring study puts a lot of emphasis on ethical responsibility by carefully thinking about important issues like privacy, justice, preventing misuse, and sustainability. The study only uses publicly available datasets that have been ethically sourced to protect privacy. This means that no personal or sensitive data is put at risk. The method reduces reliance on external facial images by creating a reference-free deblurring technique. This greatly lowers the risk of unauthorized use of people's likenesses or the possible exposure of private information. This focus on privacy goes hand in hand with a commitment to justice, as the method aims to produce fair and unbiased results, avoiding the spread of societal biases that could happen if datasets are skewed or not representative. The study suggests that the technology should only be used in safe and controlled environments where it can be monitored to make sure it is used responsibly and to lower the risks of things like creating deepfakes or spying without permission. The research also supports sustainability by using efficient algorithms that lower the amount of computing power needed, which in turn lowers energy use and environmental impact. The study promotes openness and responsibility by making its documentation available to everyone. This encourages the research community and other interested parties to carefully examine and responsibly use the study's findings. By following these ethical rules, the study not only pushes the boundaries of technical innovation, but it also sets a standard for responsible development in the area of facial image processing.

## 1.14 Thesis Management and Timeline

Figures 1.3 show a Gantt Chart showing the work breakdown of this thesis and the time-span of each breakdown subtask. The study lasted more than a year covering topic selection, literature review, goal definition, experiment design, and documentation.

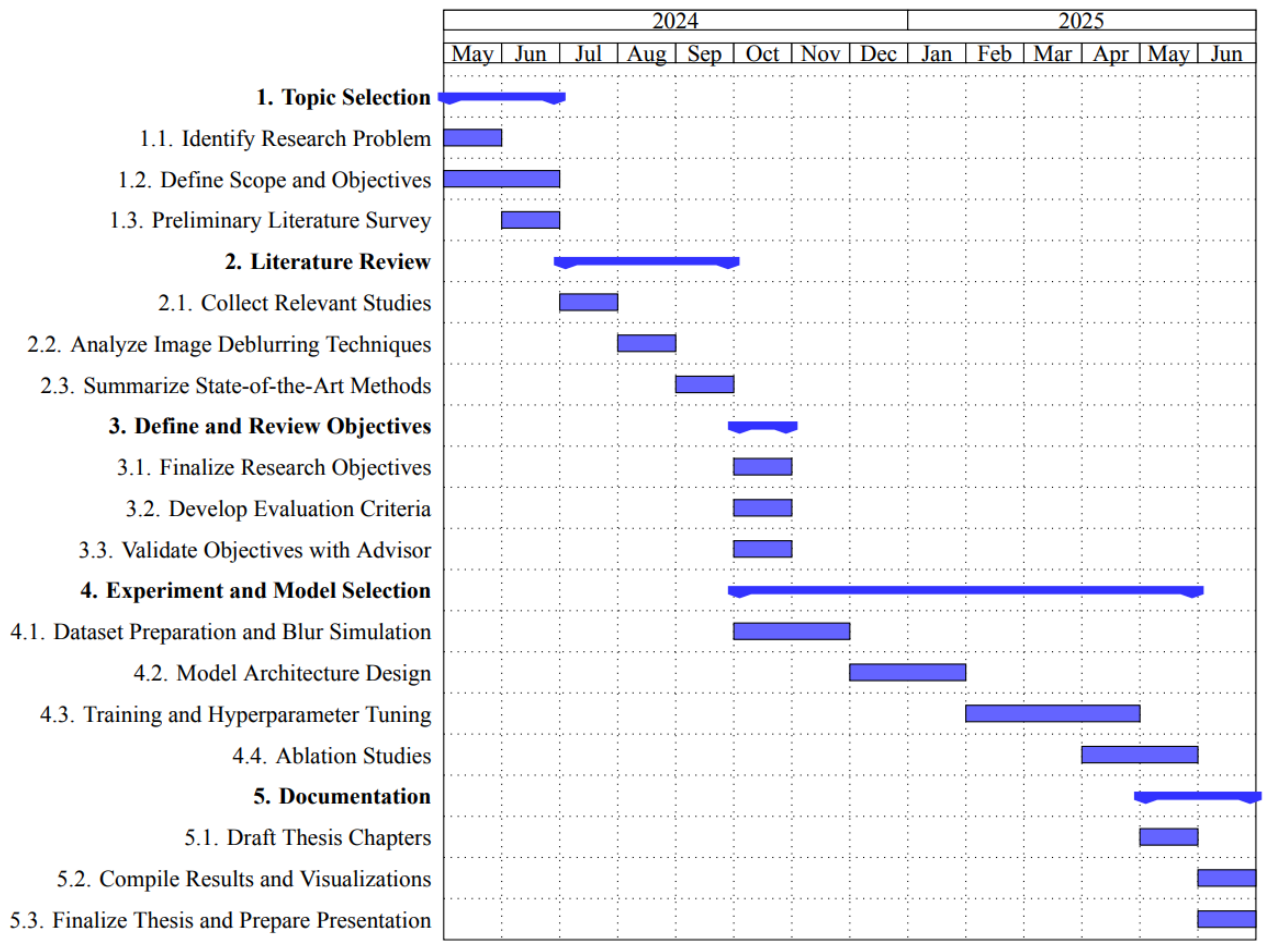

Figure 1.3: Gantt Chart with Work Breakdown Structure for Research Project Timeline

## 1.15 Thesis Organization

The latter part of this thesis book is organized as follows:

- **Chapter 2:** Describes the background studies required and reviews literature on image deblurring techniques, especially for facial images.

- **Chapter 3:** Describes the overall methodology, the dataset, blur simulation, model ar-chitecture, and experimental setup.

- **Chapter 4:** Presents experimental results, ablation studies, and comparisons with state-of-the-art methods.

- **Chapter 5:** Summarizes findings and outlines potential directions for future work.

## 1.16  Conclusion

Dedicated facial deblurring procedures are required since traditional deblurring algorithms can-not preserve the geometric and identity-specific properties required for facial images. Semantic-based approaches preserve these structures, but their reliance on reference images limits their usefulness. This paper proposes a novel method for producing semantic facial masks directly from blurry images, thereby eliminating reference image dependence and increasing robustness. The lightweight CNN architecture ensures compatibility with low-end devices, improving its real-world applicability. This chapter outlines the study concerns, aims, and challenges in de-termining whether semantic masks from blurry inputs can improve deblurring, preserve iden-tity without references, and allow lightweight models to compete with state-of-the-art systems. One of the goals is to develop a computationally efficient, reference-free, identity-preserving deblurring system capable of handling extreme blur, generating precise semantic masks, bal-ancing model complexity, and duplicating realistic blur. This sophisticated solution, which is detailed here, was developed in a single year and has applications in surveillance, forensics, and facial recognition. The following chapter provides background research and a thorough overview of the literature.

# Chapter 2

# Background Study and Literature Review

## 2.1 Introduction

This chapter defines the essential mathematical and engineering concepts required to examine face picture deblurring, including digital image processing, deep learning, and convolutional neural network modules, thereby establishing the groundwork for the research presented in this thesis. It describes the highly sophisticated modern engineering instruments employed in the investigation. A comprehensive literature review is presented, focusing on existing picture deblurring approaches, with a special emphasis on facial image deblurring, highlighting their advancements and limitations. The chapter also discusses data analysis methods. Establish-ing this theoretical and pragmatic foundation prepares the reader for the proposed approach. It provides a clear grasp of the problems and possibilities in developing effective deblurring procedures for facial photographs.

## 2.2 Required Advanced Knowledge of Mathematics and En-gineering Sciences

This work on face image deblurring requires a strong awareness of intricate mathematical ideas and technical domains. Mathematical modeling, deep learning, and digital image processing all significantly impact the methods applied, particularly in the construction of semantic masks and the development of deep CNN-based architectures. The main knowledge domains needed to grasp and implement the intended research are discussed in this part.

## 2.2.1 Digital Image Processing

A basic field encompassing methods such as convolution, filtering, edge detection, image alter-ation, and enhancement. Tasks such as picture resizing, normalizing, histogram equalization, and semantic segmentation are used to create facial structure masks for guiding deblurring mod-els, which rely on this knowledge. Digital image processing is essential to grasp the kinds of blurriness and how they affect images as well as to articulate them numerically.

## 2.2.2 Gaussian Blur

Gaussian blur models the blurry image caused by being out of focus. By convolution of the image with a Gaussian kernel, a typical image smoothing method called Gaussian blur lowers noise and fine detail. These kernels have specific shapes [18]. Synthetic blur modeling is extensively applied in assessing deblurring techniques.

The 2D Gaussian function is given by:

$$G(x, y) = \frac{1}{2\pi\sigma^2} \exp\left(-\frac{x^2 + y^2}{2\sigma^2}\right) \quad (2.1)$$

Blurring is performed by convolving this kernel with the image:

$$I_{\text{blurred}}(i, j) = \sum_m \sum_n G(m, n) \cdot I(i - m, j - n) \quad (2.2)$$

As an example, considering a 6 × 6 grayscale image:

$$I = \begin{bmatrix} 10 & 20 & 30 & 40 & 50 & 60 \\ 15 & 25 & 35 & 45 & 55 & 65 \\ 20 & 30 & 40 & 50 & 60 & 70 \\ 25 & 35 & 45 & 55 & 65 & 75 \\ 30 & 40 & 50 & 60 & 70 & 80 \\ 35 & 45 & 55 & 65 & 75 & 85 \end{bmatrix}$$

Using a simple normalized 2 × 2 Gaussian kernel:

$$G = \frac{1}{4} \begin{bmatrix} 1 & 1 \\ 1 & 1 \end{bmatrix}$$

To compute the blurred value at position (1, 1), convolve $G$ over the top-left 2×2 submatrix:

$$I_{\text{blurred}}(1,1) = \frac{1}{4}(10 + 20 + 15 + 25) = \frac{70}{4} = 17.5$$

This process is repeated for all positions, yielding a smoothed version of the original image. Gaussian blur smoothens images and this is also applied in high noisy images to reduce noise [19].

### 2.2.3 Motion Blur

Image smearing results from relative motion between the camera and the object during expo-sure producing motion blur. In reality, motion blur is not always uniform but non uniform motion blur can also be deblurred [20]. Usually modeling directional motion, it is modeled as a convolution with a linear point spread function (PSF).

Mathematically, the motion-blurred image is given by:

$$I_{\text{blurred}}(i, j) = \sum_m \sum_n K(m, n) \cdot I(i - m, j - n) \tag{2.3}$$

where $K$ is the PSF kernel representing the motion direction and length.

As an example, considering a $6 \times 6$ grayscale image:

$$I = \begin{bmatrix} 10 & 20 & 30 & 40 & 50 & 60 \\ 15 & 25 & 35 & 45 & 55 & 65 \\ 20 & 30 & 40 & 50 & 60 & 70 \\ 25 & 35 & 45 & 55 & 65 & 75 \\ 30 & 40 & 50 & 60 & 70 & 80 \\ 35 & 45 & 55 & 65 & 75 & 85 \end{bmatrix}$$

Assuming horizontal motion, we use a $1 \times 3$ PSF kernel:

$$K = \frac{1}{3} \begin{bmatrix} 1 & 1 & 1 \end{bmatrix}$$

To compute the blurred value at position $(1, 2)$:

$$I_{\text{blurred}}(1, 2) = \frac{1}{3}(20 + 30 + 40) = \frac{90}{3} = 30$$

This convolution simulates linear motion blur over the image.

Accurate synthesis of synthetic blurred data and creation of deep models capable of per-forming temporal deblurring depend on an awareness of motion blur in this form.

## 2.2.4 Deep Learning Theory

Deep learning is a subfield of machine learning whereby automatic feature extraction from data using multi-layered artificial neural networks. Unlike other conventional approaches, it can learn straight from raw inputs very complicated patterns at any number of scales and does not call for human manual feature engineering. Over the past decade, deep learning has been a crucial advancement in artificial intelligence. It excels at image processing [21].

Among the fundamental ideas are architectural elements include activation functions, nor-malizing layers, and—above all—skip connections; also include gradient descent optimization, backpropagation to update weights. Furthermore covered in the theory are generalization, over-fitting, and convergence events.

Development and training models such U-Net variations applied to picture deblurring, which can learn spatial features effectively, improve reconstruction quality, and maximize model per-formance, need on a good knowledge of deep learning.

## 2.2.5 Convolutional Neural Networks (CNNs)

Made of neurons that process inputs using dot products and non-linearities, convolutional neural networks (CNNs) share characteristics with standard neural networks. Learnable weights and biases abound among these neurons. They translate raw picture pixels to class scores using a differentiable function in the last fully-connected layer and a loss function. Unlike traditional neural networks, CNNs view inputs as images, which enables customized designs that maximize efficiency and require less variables than the network itself. Inspired by animal visual systems, CNNs learn features straight from data rather than hand-crafted ones [22]. Following several nonlinear transformation layers, each of which use feature vectors as inputs and outputs, is a supervised classifier. CNNs require large labeled datasets and significant processing capability, which GPUs and an abundance of digital data help to enable [23].

Designed for processing images with little preprocessing, a CNN is a feed-forward neural network. Arranged in three dimensions—width, height, depth—neurons have learnable weights and biases. Forward passes produce class score predictions, errors are calculated, and backprop-agation modifies parameters to repeatedly reduce mistakes throughout training. CNNs effec-tively translate 3D input volumes—e.g., RGB images—into 3D output volumes by connecting neurons to small areas unlike conventional neural networks.

A convolutional neural network (CNN) generates class scores from input data by means of convolutional, pooling, fully-connected, and activation layers. Using learnable filters, the con-volutional layer produces activation maps under control via zero-padding, stride, and depth. ReLU activation layers bring non-linearity while pooling layers—such as max pooling—or transposition layers downsample spatial dimensions to reduce computation and overfitting. Often used in transposed convolutions (or deconvolutions), transposition layers upsample or downsample by learning weights to change spatial dimensions unlike fixed pooling processes. Usually employing softmax, fully-connected layers can be interconverted with convolutional layers and compute final outputs.

### 2.2.6 Convolutions

Handling most of computing chores, the convolutional (Conv) layer forms the backbone of a convolutional neural network (CNN) [24]. It makes use of learnable filters, usually tiny in spatial dimensions (e.g., 3×3) yet encompassing the whole depth of the input volume (e.g., 3 for RGB channels). Each filter slides across the input during the forward pass to compute dot products creating a 2D activation map identifying patterns or edges. Multiple filters—say, eight—create stacked activation maps that generate a three-dimensional output volume. Three hyperparameters—depth, number of filters, stride—step size of filter movement—determine the output size: zero-padding, border padding, depth Calculating the output's spatial size results in:

$$\text{Output size} = \left\lfloor \frac{W - F + 2P}{S} \right\rfloor + 1,$$

(2.4) where $W$ is the input size, $F$ is the filter size, $P$ is the padding, and $S$ is the stride.

To illustrate, consider a 5×5 grayscale image (depth=1) as the input:

$$\begin{bmatrix} 1 & 2 & 3 & 4 & 5 \\ 6 & 7 & 8 & 9 & 10 \\ 11 & 12 & 13 & 14 & 15 \\ 16 & 17 & 18 & 19 & 20 \\ 21 & 22 & 23 & 24 & 25 \end{bmatrix}$$

Using a 3×3 filter:
$$\begin{bmatrix} 1 & 0 & -1 \\ 2 & 0 & -2 \\ 1 & 0 & -1 \end{bmatrix}$$

With stride=1 and no padding ($P = 0$), the output size is:
$$\left\lfloor \frac{5 - 3 + 2 \cdot 0}{1} \right\rfloor + 1 = 3.$$

The dot product for the top-left 3×3 region is:

$$(1 \cdot 1) + (2 \cdot 0) + (3 \cdot -1) + (6 \cdot 2) + (7 \cdot 0) + (8 \cdot -2) + (11 \cdot 1) + (12 \cdot 0) + (13 \cdot -1) = -8.$$

Sliding the filter across all positions yields a 3×3 activation map (simplified for illustration).

For the same input with padding=1 (zeros around the border) and stride=2, the padded input becomes 7×7. The output size is:

$$\left\lfloor \frac{7 - 3 + 2 \cdot 1}{2} \right\rfloor + 1 = 4.$$

For the top-left region:

$$\begin{bmatrix} 0 & 0 & 0 \\ 0 & 1 & 2 \\ 0 & 6 & 7 \end{bmatrix}$$

The dot product is:

$$(0 \cdot 1) + (0 \cdot 0) + (0 \cdot -1) + (0 \cdot 2) + (1 \cdot 0) + (2 \cdot -2) + (0 \cdot 1) + (6 \cdot 0) + (7 \cdot -1) = -11.$$

Moving with stride of two, the filter generates a 4× 4 activation map. These scenarios show how conv layers extract features under hyperparameter control of output dimensions.

### 2.2.7 Dense Convolution

Within convolutional neural networks, a dense convolution block [25] acts as a module creating direct connections among all layers with compatible feature-map dimensions. First suggested within the DenseNet framework was this idea. Every layer combines extra inputs from all previous layers and sends its feature-maps to all next layers, therefore preserving the feedfor-ward process. Dense Blocks use concatenation unlike ResNets, which aggregate characteristics before layer input by means of summation. The $\ell$ layer thus manages $\ell$ inputs, including feature-maps from all previous convolutional blocks, and outputs are provided to the remaining $L - \ell$ layers. Unlike conventional topologies, this arrangement produces $\frac{L(L+1)}{2}$ connections over a $L$-layer network, hence introducing the idea of "dense connectivity". A dense block is seen in Figure 2.1.

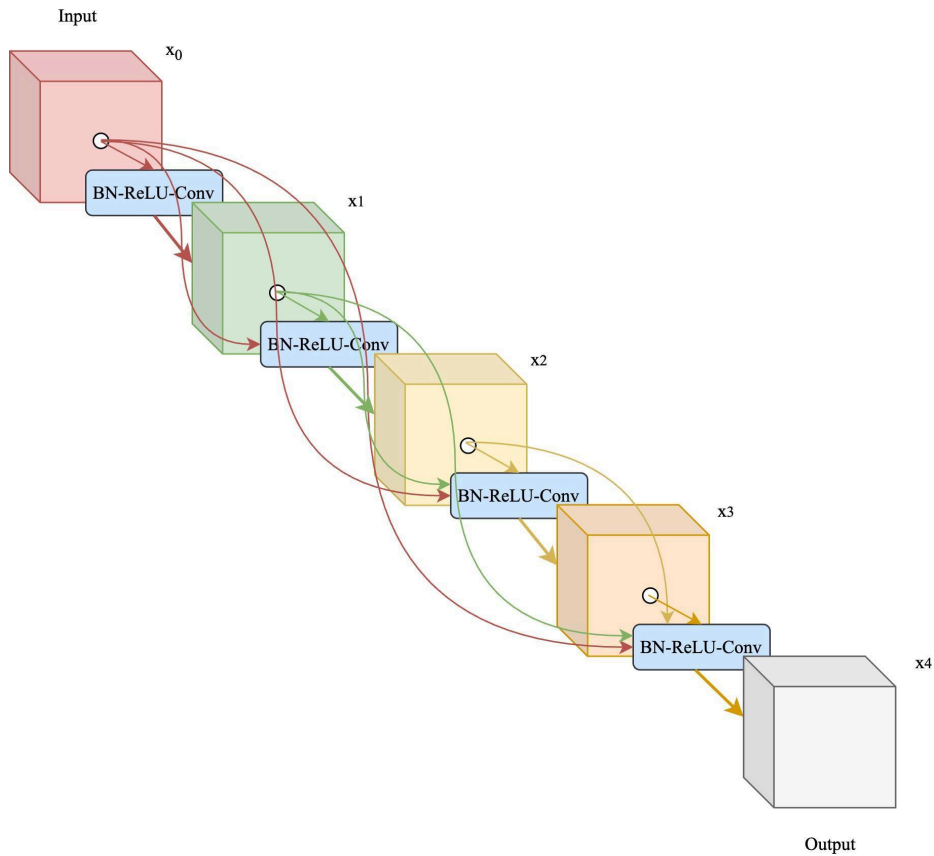

Figure 2.1: A dense block, where each layer utilizes all preceding feature-maps as input.

### 2.2.8 Upsampling in CNN

When the output of convolutional neural networks (CNNs) must match the input size—as in semantic segmentation tasks where every input pixel is allocated a label—upsampling is ab-solutely essential. Though not the precise feature maps, this approach reverse down sampling to recover the original spatial dimensions. Aimed toward reconstruction of the original image size, upsampling is a fundamental component of the later part of a fully convolutional network.

### 2.2.9 Simple Upsampling (Nearest Neighbor)

A fundamental and computationally light approach is simple upsampling, sometimes known as Nearest Neighbour interpolation. Repeating rows and columns depending on an upsampling factor raises the spatial resolution. For a factor of (2, 2), for instance, each pixel is doubled in width and height.

Nearest neighbour pooling is exemplified here.

$$\begin{bmatrix} 1 & 2 \\ 3 & 4 \end{bmatrix} \rightarrow \begin{bmatrix} 1 & 1 & 2 & 2 \\ 1 & 1 & 2 & 2 \\ 3 & 3 & 4 & 4 \\ 3 & 3 & 4 & 4 \end{bmatrix}$$

$$1\ 2 \quad \rightarrow \quad [1\ 1\ 2\ 2]$$

### 2.2.10 Un-pooling

In convolutional neural networks (CNNs), un-pooling [26] is a method used to reconstruct a higher-order feature map so reversing the downsampling effect of pooling layers, including max pooling. By choosing the maximum value within a window (e.g., 2×2) with a defined stride, max pooling decreases spatial dimensions while preserving positional information of these maxima, therefore acting as the "switch". Un-pooling fills in remaining areas with zeros by re-positioning these values in an upsampled grid. Information loss during pooling prevents this technique from recovering the original data even though it is efficient at restoring resolution.

Here is an example of a 2×2 matrix resulting from max pooling:

$$\begin{bmatrix} 5 & 8 \\ 3 & 7 \end{bmatrix}$$

Assuming a 4×4 input was pooled with a 2×2 window and stride 2, un-pooling reconstructs a 4×4 grid. Steps include: (1) initializing a zero-filled 4×4 matrix, and (2) placing max values (5, 8, 3, 7) at their original positions (e.g., (1,1), (1,4), (3,1), (3,4)) based on the switch.

$$\begin{bmatrix} 5 & 8 \\ 3 & 7 \end{bmatrix} \rightarrow \begin{bmatrix} 0 & 0 & 0 & 0 \\ 0 & 0 & 0 & 0 \\ 0 & 0 & 0 & 0 \\ 0 & 0 & 0 & 0 \end{bmatrix} \rightarrow \begin{bmatrix} 5 & 0 & 0 & 8 \\ 0 & 0 & 0 & 0 \\ 3 & 0 & 0 & 7 \\ 0 & 0 & 0 & 0 \end{bmatrix}$$

### 2.2.11 Transpose Convolution

Often called deconvolution, transpose convolution maps a smaller input to a bigger output so reversing the spatial reduction of conventional convolution [27]. For every input element it uses a learnable kernel to produce an output grid with stride-defined spacing. Commonly used in generative models and semantic segmentation, overlapping areas are summed to allow trainable upsampling.

Here is an example for a 2×2 input and 2×2 kernel:

$$\text{Input: } \begin{bmatrix} 1 & 2 \\ 3 & 4 \end{bmatrix}, \quad \text{Kernel: } \begin{bmatrix} 1 & 2 \\ 3 & 4 \end{bmatrix}$$

With stride 2, the output is a 3×3 matrix. Steps include: (1) initializing a 3×3 zero matrix, and (2) applying the kernel to each input element, placing results at stride-adjusted positions, and summing overlaps. Step-by-step operation is shown below:

$$\begin{bmatrix} 1 & 2 \\ 3 & 4 \end{bmatrix} \rightarrow \begin{bmatrix} 0 & 0 & 0 \\ 0 & 0 & 0 \\ 0 & 0 & 0 \end{bmatrix} \rightarrow \begin{bmatrix} 1 & 2 & 0 \\ 3 & 4 & 0 \\ 0 & 0 & 0 \end{bmatrix} \rightarrow \begin{bmatrix} 1 & 4 & 4 \\ 3 & 10 & 8 \\ 0 & 0 & 0 \end{bmatrix} \rightarrow \begin{bmatrix} 1 & 4 & 4 \\ 3 & 10 & 8 \\ 9 & 6 & 0 \end{bmatrix} \rightarrow \begin{bmatrix} 1 & 2 & 4 \\ 3 & 10 & 8 \\ 9 & 12 & 16 \end{bmatrix}$$

### 2.2.12 Pixel Shuffle Upsampling

An effective method extensively utilized in super-resolution and deblurring applications is pixel shuffle upsampling [28], sometimes known as sub-pixel convolution. It converts items from the channel dimension of a feature map into spatial dimensions, therefore improving resolution without adding checkerboard artifacts typical of transpose convolution. Pixel shuffle reshapes an input of size $H \times W \times C \cdot r^2$, where $r$ is the upsampling factor, so dispersing channel infor-mation into a higher-order spatial grid. An instance of the Pixel Shuffle Upsampling process of an input of 3x3 with four channels producing an output of 6x6 with one channel is shown in Figure 2.2.

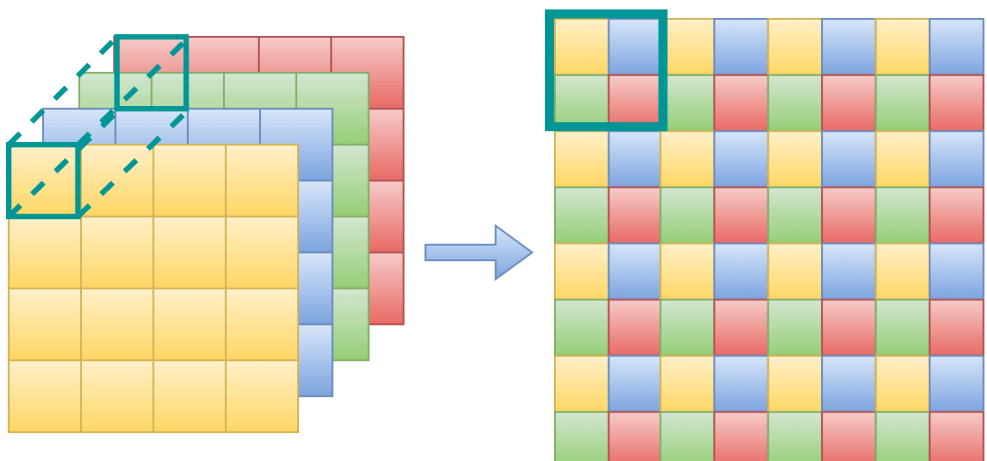

Figure 2.2: Pixel Shuffle Upsamping of an Input of four Channels

## 2.2.13 Downsampling

In convolutional neural networks (CNNs), downsampling lowers the spatial dimensions of fea-ture maps following first feature extraction, hence minimizing computational complexity while maintaining important information. Computationally costly is direct feeding high-dimensional outputs from the first convolutional layer to further layers. Usually done in the first half of fully convolutional networks, downsampling reduces the feature map size with least effect on retrieved features.

The output dimensions of a convolutional layer are determined by:

$$\text{new\_rows, new\_cols} = \frac{\lfloor n + 2p - f \rfloor}{s} + 1$$

where $n$ is the input height or width, $p$ is padding, $f$ is filter size, and $s$ is stride. The full output shape is:

$$\text{batch\_shape} + (\text{new\_rows, new\_cols, no. of filters})$$

Increasing the stride ($s > 1$) reduces the output size, facilitating downsampling.

CNNs make two main downsampling decisions using pooling and strided convolution. Skipping input points in convolution with a stride larger than one lowers spatial dimension-ality. For instance, a 2x 2 filter running stride 2 drastically reduces the output size. One finds extensive application in pooling, especially max pooling. With a designated stride, it covers the input a window (e.g., 2x2) choosing the maximum value (or average for average pooling). For instance, max pooling with pool size (2,2) and stride 2 decreases the input size to 25% of its original dimensions. Another type of pooling is average pooling. An illustration of Max Pooling and Average Pooling operation is provided by Figure 2.3.

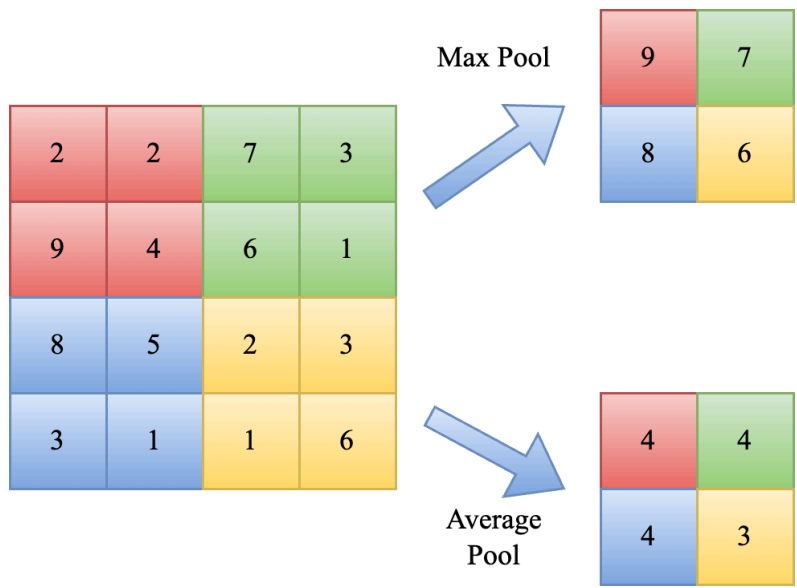

Figure 2.3: Max and Average Pooling on a 4x4 Input with Stride 2

## 2.2.14 Attention Mechanism

Inspired by human visual selective focus, attention mechanisms in convolutional neural net-works (CNNs) improve performance by letting models concentrate on pertinent areas of input data [29]. By giving discriminative features top priority and noise reduction top importance, they solve limits of conventional CNNs—which treat all picture areas identically. Among the main advantages are better interpretability, strong resistance to complicated data, and improved feature extraction. Operating by creating attention maps that weight feature maps, attention mechanisms recalibrate them to highlight key areas or channels. Common forms are chan-nel attention (e.g., Squeeze-and-Excitation block [30]), which represents inter-channel depen-dencies, spatial attention (e.g., PSANet [31]), which focuses on geographic areas, and hybrid attention (e.g., CBAM [32]), thereby aggregating both. For tasks such as image classifica-tion, object detection, and segmentation—where they increase accuracy by collecting contex-tual relationships—these mechanisms are essential. Notwithstanding their benefits, problems include computing overhead and guaranteeing generalization over tasks.

## 2.2.15 Convolutional Block Attention Module (CBAM)

By progressively applying channel and spatial attention, the Convolutional Block Attention Module (CBAM) [32] refines feature maps. Global average and max pooling are used to col-lect spatial information for channel attention with which dense layers with ReLU and sigmoid activations produce channel weights $\mathbf{W}_c$. $\mathbf{W}_c$ scales the input feature map $x$ to underline signif-icant channels: $x \otimes \mathbf{W}_c$ Average and max pooled features concatenated and processed through a convolutional layer with sigmoid activation create spatial weights $\mathbf{W}_s$ for spatial attention.

To concentrate on significant spatial areas, the feature map $x_c$ is subsequently scaled by $\mathbf{W}_s$ $x_{\text{out}} = x_c \otimes \mathbf{W}_s$ Combining channel and spatial attention, CBAM improves feature representa-tions and so is useful for tasks needing exact feature localization. Figure 2.4 shows an abstract concept of CBAM attention.

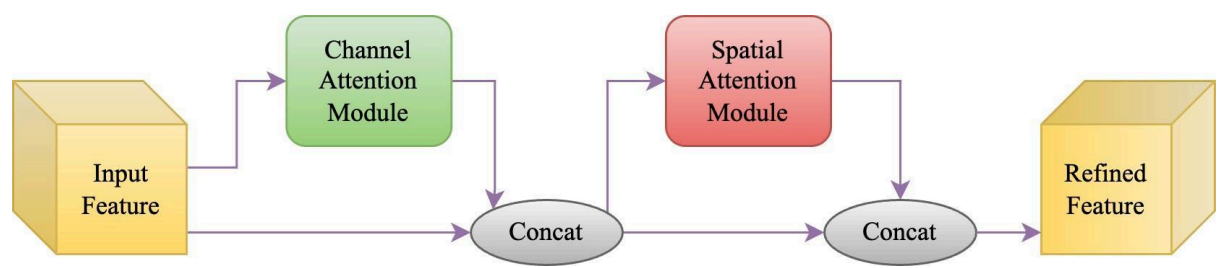

Figure 2.4: The Convolutional Block Attention Module

## 2.2.16 Batch Normalization

By averaging mini-batch activations to stabilize layer input distributions during training, batch normalization [33] solves internal covariate shift in neural networks. For a mini-batch of activations $\{x_1, x_2, \ldots, x_m\}$, it computes the mean $\mu_B = \frac{1}{m} \sum_{i=1}^{m} x_i$ and variance $\sigma_B^2 = \frac{1}{m} \sum_{i=1}^{m} (x_i - \mu_B)^2$. Each activation is normalized as: $\hat{x}_i = \frac{x_i - \mu_B}{\sqrt{\sigma_B^2 + \epsilon}}$ where $\epsilon$ is a small constant for numerical stability. Normalized activations are then scaled and shifted using learnable parameters $\gamma$ and $\beta$: $y_i = \gamma \hat{x}_i + \beta$ This strategy guarantees consistent training, speeds convergence, promotes faster learning rates, minimizes vanishing or exploding gradients, and offers a regularizing impact, therefore lessening the need for dropout techniques. It improves work performance including picture classification.

## 2.2.17 Residual Connection

Introduced in ResNet [34], residual connections enable layers learn the *residual* $F(x) = H(x) - x$ instead of the whole transformation $H(x)$. This helps to simplify deep network training. Learning $H(x) = x$ (identity) for instance is difficult; its residual $F(x) = 0$ is simple. The result comes to be $F(x) + x = H(x)$. Shortcuts in Residual Connection help to improve gradient flow in deep networks. Residual blocks iteratively polish features; more blocks improve accuracy. An example of a residual connection is shown in figure 2.5.

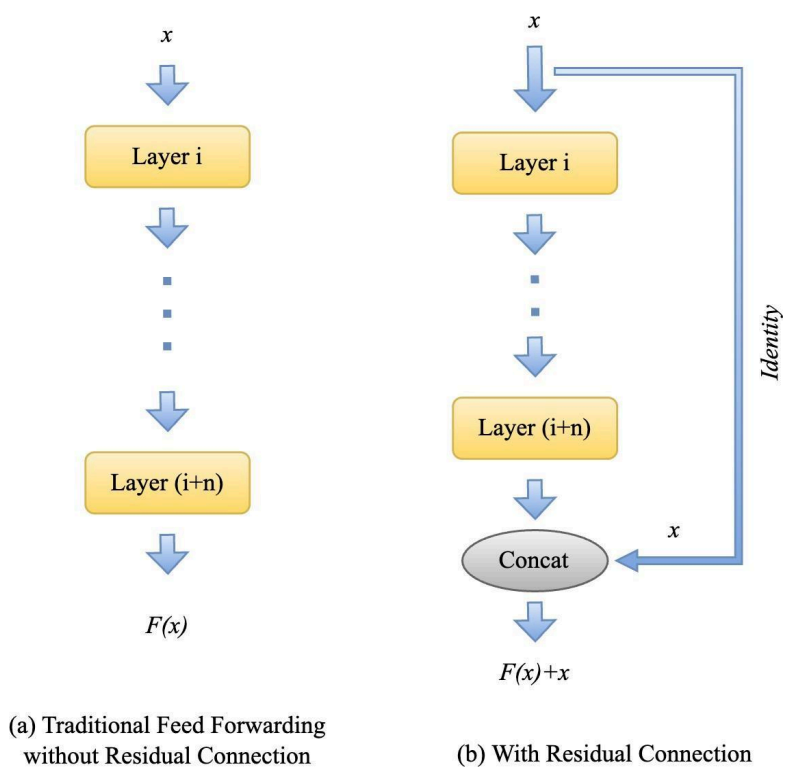

Figure 2.5: A traditional feed forwarding without residual connection vs with residual connec-tion

## 2.2.18 Activation Functions

Mathematical procedures applied to neuron outputs, activation functions create non-linearity and let neural networks learn intricate patterns [35]. Networks devoid of non-linearity would function as linear regression, confined to straightforward, linearly separable challenges.

To find neuron activation, build curved decision boundaries for complicated data, and offer gradients for backpropagation to change weights and biases, activation functions compute the weighted sum of inputs plus bias. Introduced by ReLU, Sigmoid, or Tanh, non-linearity lets one model non-linearly separable data and supports advanced deep learning projects. Figure 2.6 displays in a neural network where and how the activation function runs.

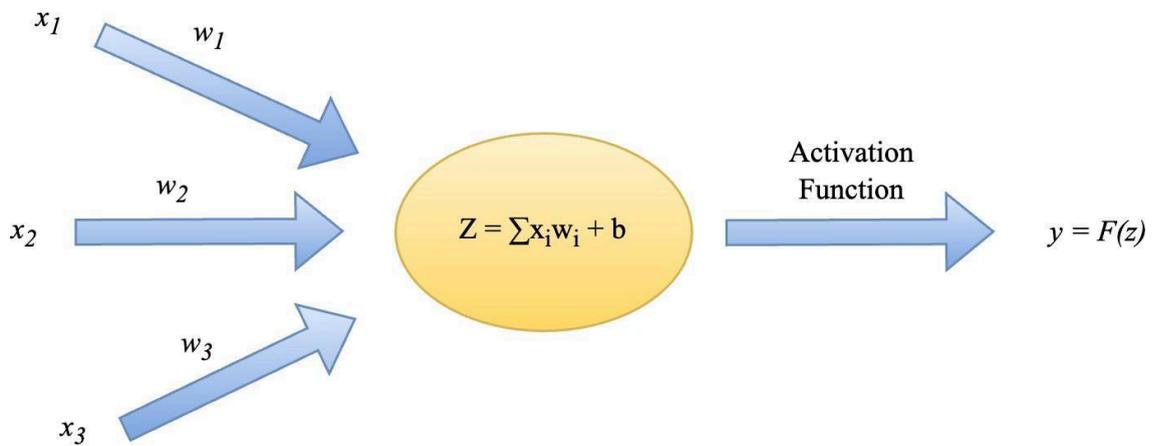

Figure 2.6: Activation Function in a Neural Network

A common activation function is ReLU:

$$\sigma(x) = \max(0, x)$$

Consider a neural network with two inputs $i_1$, $i_2$, a hidden layer with neurons $h_1$, $h_2$, and one output neuron. The hidden layer computes:

$$h_1 = i_1 w_1 + i_2 w_3 + b_1$$

$$h_2 = i_1 w_2 + i_2 w_4 + b_2$$

A linear output is:

$$\text{output} = h_1 w_5 + h_2 w_6 + \text{bias}$$

Applying the Sigmoid activation function:

$$\sigma(x) = \frac{1}{1 + e^{-x}}$$

$$\text{final output} = \sigma(h_1 w_5 + h_2 w_6 + \text{bias}) = \frac{1}{1 + e^{-(h_1 w_5 + h_2 w_6 + \text{bias})}}$$

This non-linearity lets the network pick out complex patterns. Absence of non-linearity required for complex pattern learning, a linear activation function generates an output exactly proportional to the input. For instance, $(x) = x$. This linear activation function is shown by figure 2.7.

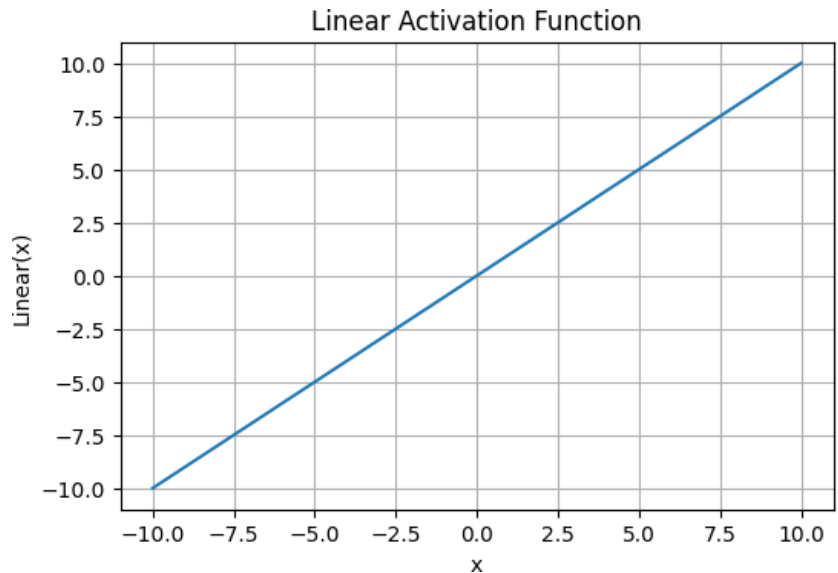

Figure 2.7: Linear Activation Function

Curvature introduced by non-linear functions helps neural networks to represent intricate, non-linearly separable data. ReLU, for instance, has $\sigma(x) = \max(0, x)$.

### 2.2.19 Sigmoid Activation Function

The Sigmoid function, a non-linear activation function, characterized by its 'S' shape, is defined as:

$$\sigma(x) = \frac{1}{1 + e^{-x}}$$

It produces a smooth, continuous output between 0 and 1, ideal for gradient-based optimization. The output range of the function—0 to 1—makes it appropriate for binary categorization. For inputs between -2 and 2, it shows a steep gradient where little input changes produce sub-stantial output alterations, hence supporting training sensitivity. For great input magnitudes, it can, however, suffer from disappearing gradients.

For binary classification problems, including probability prediction—that is, spam against not spam—sigmoid is commonly utilized in output layers. Although its smooth gradient fa-cilitates steady training, gradient problems make it less frequent in hidden layers. This linear activation function is demonstrated by figure 2.8.

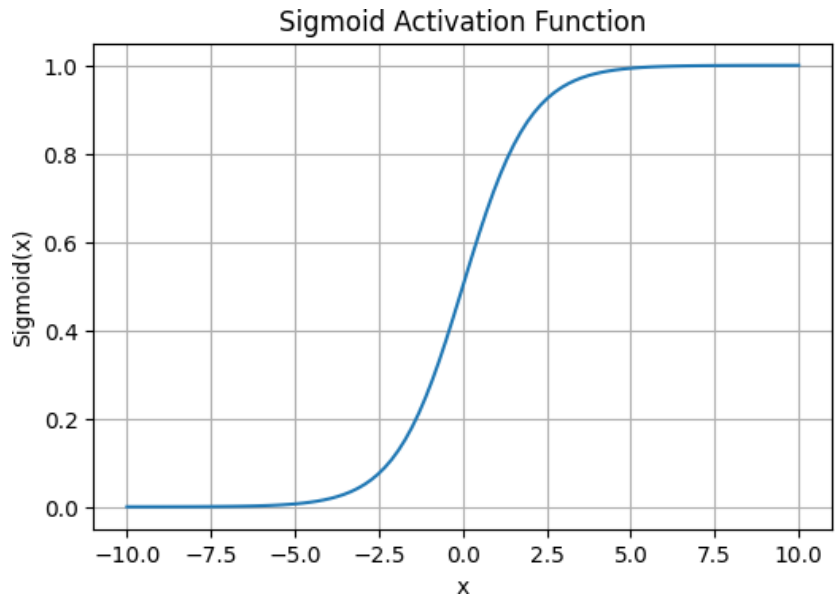

Figure 2.8: Sigmoid Activation Function

### 2.2.20 Tanh Activation Function

The hyperbolic tangent (Tanh), a non-linear activation function, function is defined as:

$$\tanh(x) = \frac{2}{1 + e^{-2x}} - 1$$

Alternatively, it can be expressed as:

$$\tanh(x) = 2\sigma(2x) - 1$$

where $\sigma$ is the Sigmoid function. It outputs values between -1 and 1.

Because tanh is zero-centered and non-linear, its outputs are symmetric about zero, which speeds up learning in next layers. Although its gradient is greater than Sigmoid's, severe inputs still run the danger of having vanished slopes.

Because of its zero-centered output, tanh is widely utilized in hidden layers of neural networks—especially in recurrent neural networks (RNNs). It models difficult patterns in applications in-cluding natural language processing and time-series prediction. This tanh activation function is displayed by figure 2.9.

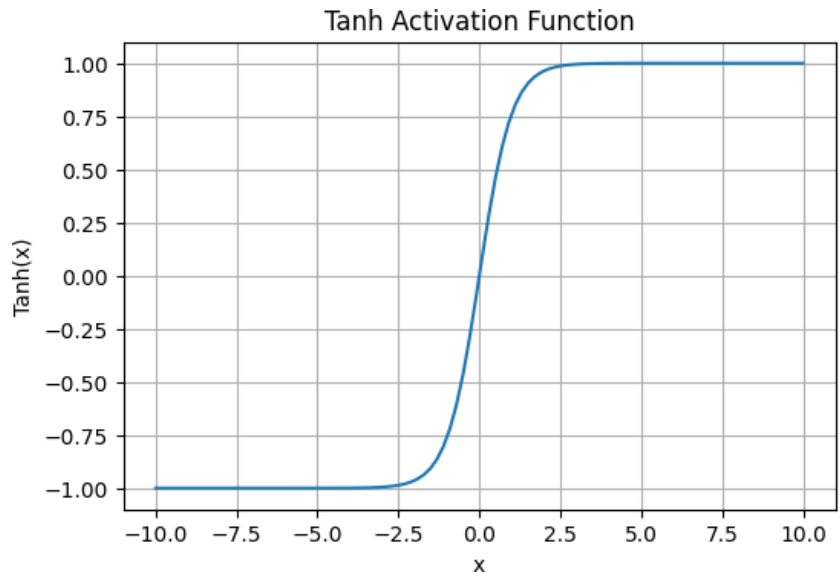

Figure 2.9: Tanh Activation Function

## 2.2.21 ReLU Activation Function

Another non-linear activation function is ReLU, first used in CNN by Krizhevsky et. al. [22], Rectified Linear Unit, is defined as:

$$\sigma(x) = \max(0, x)$$

It outputs $x$ if positive; otherwise, it returns 0, creating a simple threshold.
ReLU is non-linear, with an output range of $[0, \infty)$. Its simplicity reduces computational cost compared to Sigmoid and Tanh, and it mitigates vanishing gradients by maintaining a constant gradient of 1 for positive inputs. However, it can suffer from "dying ReLU" where neurons output zero permanently.

Particularly in convolutional neural networks (CNNs) for image processing and deep learn-ing models, ReLU finds extensive application in hidden layers of deep neural networks. Large-scale projects like object detection would find it perfect since of its efficiency and capacity to understand intricate patterns. ReLU activation function is demonstrated by figure 2.10.

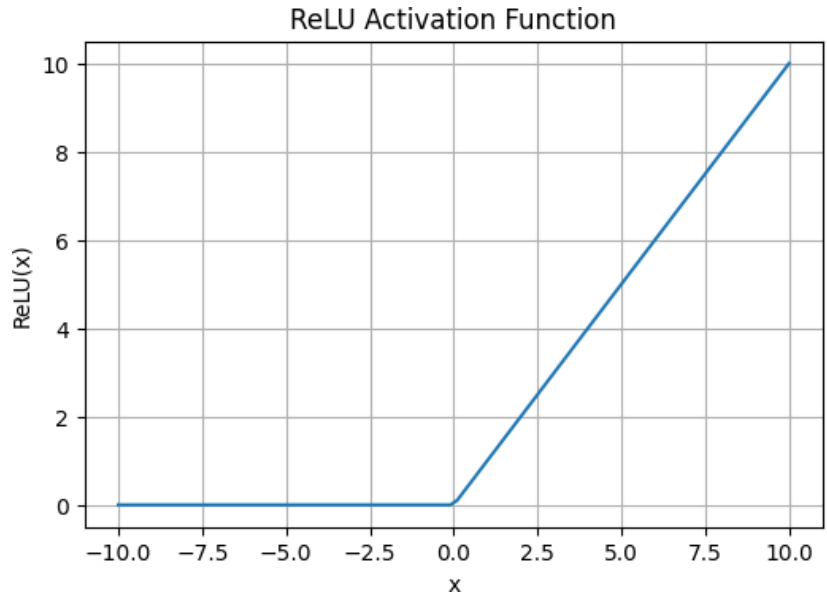

Figure 2.10: ReLU Activation Function

## 2.2.22 Adam Optimizer

The Adam [36] optimizer, Adaptive Moment Estimation, is a popular optimization approach used in deep neural network training. It combines the benefits of two current techniques: RM-SProp, which adjusts learning rates for each parameter, and momentum, which accelerates gra-dients in the desired direction. Adam can acquire adaptive learning rates by keeping two moving averages: the first instant (mean) of the gradients and the second moment (uncentered variance). This allows for faster convergence and great robustness to noisy gradients.

Adam is widely selected due to its ease of implementation, low memory requirements, and efficiency. Default hyperparameter values (e.g., learning rate = 0.001, β = 0.9, β = 0.999) are typically sufficient for most applications and require minimal change. On some situations, however, generalization may be problematic when compared to simpler optimizers such as SGD with momentum. Weight decay variants, such as AdamW, can assist overcome some of these issues.

Adam is a practical approach for optimizing deep learning models in computer vision, nat-ural language processing, and other fields, achieving a desirable balance of speed and stability.

## 2.2.23 UNet

Designed for picture segmentation, U-Net [37] is a convolutional neural network especially useful in medical imaging for jobs including tumor identification in scans. Its U-shaped archi-tecture consists in a bottleneck storing compressed information, an expansive path (decoder) reconstructing the image using upsampling and skip connections to preserve spatial details, and a contracting path (encoder) extracting abstract features via convolutions and max pool-ing. ReLU activation brings non-linearity, which helps the model to learn intricate patterns effectively—even with insufficient labeled data—by means of which U-Net is flexible for uses like object identification and image enhancement. Figure 2.11 displays UNet's architecture.

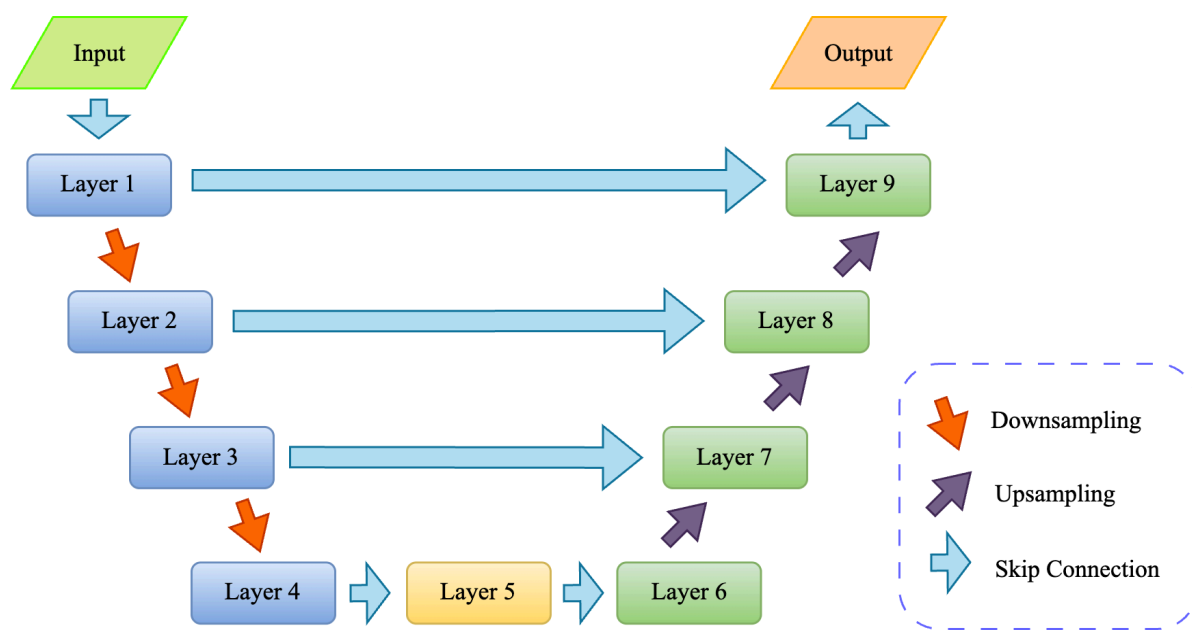

Figure 2.11: The Architecture of UNet

## 2.2.24 Multi-stage Fusion

In neural networks, multi-stage fusion is the method whereby features from several layers or stages are aggregated at several locations to improve model performance. It can combine dis-parate data streams (of different dimensionality, resolution, type, etc.) to generate information in a form that is more understandable or usable [38]. It combines low-level details (e.g., edges, textures) from early layers and high-level semantic elements (e.g., object context) from higher levels across several degrees of abstraction. Usually driven at several phases of the network, this fusion takes place via concatenation, addition, or attention

mechanisms. Multi-stage fusion

enhances the capacity of the model to capture complex patterns and dependencies by iteratively combining features, so improving its performance for tasks including image segmentation, ob-ject detection, and multi-modal learning, in which accurate predictions depend on different feature representations. One can see multi-stage fusion in Figure 2.12.

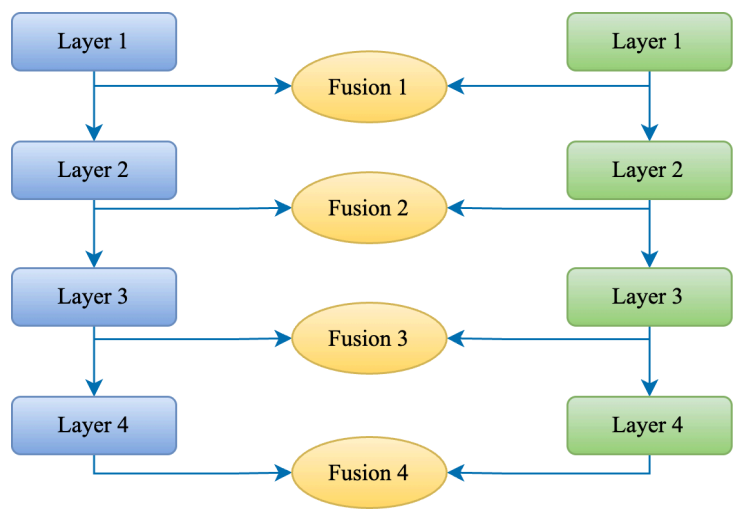

Figure 2.12: An Example of Multi-Stage Fusion

## 2.3 Utilized Modern Engineering Tools

Using Python as the main programming language, this study took advantage of libraries like NumPy for numerical operations and image processing chores, including data preparation, model creation, and evaluation. Supported convolutional layers, attention modules, and op-timization algorithms, deep learning frameworks—more especially, TensorFlow—were devel-oped, trained, and evaluated for the Semantic Mask Generator and SMFD-UNet models. Image processing tasks, including Gaussian and motion blur, image resizing, and image conversion to grayscale for semantic mask construction, were handled by OpenCV. Given mathematical mod-eling of convolution operations, OpenCV and other related tools were optional for prototyping the randomized blurring pipeline or studying blur kernel effects. Written in LaTeX, the thesis paper included pictures for technique visualization and mathematical equations, such as con-volution and post-processing calculations. Kaggle's GPU P100 was used to achieve hardware acceleration, therefore enabling effective training and inference of deep learning models, han-dling the extensive computations needed for this research. Figures containing technique flow, mask processing, and model architecture diagrams were produced using visualization tools, including Matplotlib, thereby improving the presentation of the research results in the thesis.

## 2.4 Literature Review

In computer vision, picture deblurring is a crucial chore meant to restore clear, high-quality im-ages from blurriness of inputs. Although overall picture deblurring techniques have advanced significantly, they frequently struggle with facial images because of the special difficulties pre-sented by facial structures, textures, and the need of identity preservation.

### 2.4.1 General Image Deblurring Progress and Limitations

General picture deblurring techniques seek to restore clear images from blurred inputs without particular image content restriction. One may generally classify these techniques into modern deep learning-based systems and conventional optimization-based ones.

**Traditional Methods:** Early methods based on maximum a posteriori (MAP) estimate de-pend on handcrafted priors (e.g., sparsity or gradient-based priors) to regularize the ill-posed deblurring problem. Although these techniques have limited modeling capacity, which often results in their failure to handle complicated, real-world blur even if they are good for basic blur patterns.

**Deep Learning-Based Methods:** The arrival of deep learning has transformed image de-blurring. Widely used to learn intricate blur patterns and recover high-quality photos are Gen-erative Adversarial Networks (GANs) and Convolutional Neural Networks (CNNs). Kupyn et. al. [39] for example restored clear pictures using Wasserstein GAN with perceptual loss, so sur-passing rivals in structural similarity (SSIM) and visual quality. Reiterating their earlier work, Kupyn et. al. [40] presented a relativistic conditional GAN with a double-scale discriminator, producing state-of-the-art results. Similarly, Tao et al. [41] suggested a scale-recurrent network (SRN) that processes pictures across several scales, hence greatly enhancing deblurring perfor-mance. MIMO-UNet, a single U-Net architecture with multi-scale inputs and outputs, was first presented by ChoSJ et. al. [42] which lowers running time while nevertheless preserving ex-cellent accuracy. With a lightweight architecture with hybrid feature extraction blocks, Wu et. al. [43] achieved competitive results with lowered computational costs.

Notwithstanding these developments, many times general image deblurring techniques fail on facial pictures. This is mostly resulting from the following:

- **Texture Scarcity**: Facial images often lack the rich textures and edges that general de-blurring methods depend on for kernel estimation and image restoration.

- **Identity Preservation**: Restoring facial images requires preserving fine-grained details (e.g., eyes, mouth) and maintaining the subject's identity, which is not a priority in general image deblurring.

- **Structural Regularity**: Facial images exhibit a highly regular structure, but general methods do not explicitly utilize this prior knowledge, resulting in suboptimal outcomes.

### 2.4.2 Facial Image Deblurring: Specialized Techniques

Understanding the limits of conventional deblurring techniques, scientists have created specific approaches for facial picture deblurring. These techniques get outstanding results by using face priors, semantic information, and identity-preserving algorithms.

**GAN & Diffusion Based Methods:** Because GANs can create realistic and high-quality images, they have become rather popular for facial deblurring. Employing generative facial priors from pretrained face GANs to reconstruct high-quality facial images from low-quality inputs, Wang et al. [9] presented GFP-GAN. Using diffusion models, Varanka et al. [10] sug-gested a customized face restoration approach guaranteeing faithful identity retention with fine-grained details.

**Transformer-Based Approaches:** Transformers have become effective tools for facial deblurring since they provide strong feature fusion and global context modeling. Using a reconstruction-oriented dictionary, Wang et. al. [11] generated high-quality priors, so improv-ing the restoration quality. Another improvement of their earlier work, a multi-scale multi-head cross-attention (MHCA) method to combine deteriorated features with high-quality priors, hence obtaining exceptional restoration quality, was presented by Wang et. al. [12]. Compa-rably, Zhou et. al. [13] suggested CodeFormer, a Transformer-based approach employing a discrete codebook prior to lower mapping uncertainty.

**Blind Face Restoration with Priors:** High-quality priors are used in blind face restoration techniques to retrieve information from badly deteriorated photos. Xie et al. [44] proposed PLTrans, which generates degradation-unaware representations by use of a latent diffusion-based regularization module. DFDNet, a deep face dictionary network using multi-scale com-ponent dictionaries to restore realistic details without depending on identity-specific references, was first shown in [14].

**Uncertainty-Guided Methods:** Using a confidence-guided loss to address class imbalance,

Yasarla et. al. [45] presented UMSN, a multi-stream CNN architecture that independently handles semantic classes. Without GANs, this method shown better face recognition accuracy and generated crisper images.

**Semantic-Guided Approaches:** Semantic information has shown promise in maintaining facial features and details when included into deblurring networks. Using semantic labels as both global priors and local constraints inside a multi-scale CNN, Shen et. al. [15] obtained excellent restoration with low artifact count. Likewise, Pan et. al. [16] used exemplar-based face structures to drive kernel estimate, therefore avoiding the constraints of generic priors. Using fine-tuned face parsing networks and innovative components including Facial Adaptive Denormalization (FAD) and Laplace Depth-wise Separable Convolution (LDConv) to enhance structural and texture details, Shi et. al. [17] presented RSETNet

Though much has been achieved, some issues still surround face image deblurring. Degra-dations of great degree, such severe blur or occlusions, remain challenging.

## 2.5 Techniques Utilized for Data Analysis

Techniques utilized for data analysis in this study include a combination of **quantitative**, and **qualitative** approaches to assess the performance and behavior of the proposed model.

### 2.5.1 Quantitative Analysis Techniques

A suite of measurement techniques was used to assess the quantitative analysis of produced data, each with an eye toward certain facets of data integrity and perceptual quality. These metrics included **Dice Coefficient** [46], **Dice Loss** [47], **Jaccard Index** [48], **PSNR (Peak Signal-to-Noise Ratio)** [49], **SSIM (Structural Similarity Index Measure)** [50], **MSE (Mean Squared Error)**, **NIQE (Natural Image Quality Evaluator)** [51], **LPIPS (Learned Perceptual Image Patch Similarity)** [52], and **FID (Fréchet Inception Distance)** [53], which are detailed below with their mathematical formulations and significance.

- **Dice Coefficient**: Commonly employed in image segmentation tasks to assess the similarity between predicted and ground-truth regions, dice gauges the overlap between two segmenta-tion masks. Values lie on 0 to 1; 1 denotes perfect overlap. The formula is:

$$\text{Dice} = \frac{2|A \cap B|}{|A| + |B|}$$

where *A* is the set of pixels in the predicted segmentation mask, *B* is the set of pixels in the ground-truth mask, and $|A \cap B|$ is the size of their intersection. DICE emphasizes overlap accuracy and is particularly effective for binary segmentation tasks, but it may be sensitive to small misalignments.

- **Dice Loss:** Dice loss is often used as a loss function in most of the segmentation tasks. The definition of Dice Loss is:

$$\text{Dice Loss} = 1 - \text{Dice}$$

- **Jaccard Index**: Measuring the intersection to union ratio of two segmentation masks, the Jaccard Index—also known as the Intersection over Union—IoU—quantifies their similarity. Values span 0 to 1; 1 denotes ideal overlap. The formula is:

$$\text{Jaccard} = \frac{|A \cap B|}{|A \cup B|}$$

where *A* and *B* are the predicted and ground-truth segmentation masks, respectively, $|A \cap B|$ is the size of their intersection, and $|A \cup B|$ is the size of their union. The Jaccard Index is closely related to DICE but penalizes errors more strictly, making it a robust metric for evaluating segmentation accuracy.

- **Mean Squared Error (MSE)**: A basic gauge of pixel-level accuracy, MSE computes the average squared difference between generated and reference data. Reduced MSE denotes more authenticity.

- **Peak Signal-to-Noise Ratio (PSNR)**: PSNR measures the quality of generated data by com-paring the maximum possible signal power to the noise introduced by distortions, expressed in decibels (dB).

- **Structural Similarity Index Measure (SSIM)**: Aligning with human visual perception, SSIM assesses perceptual similarity by means of brightness, contrast, and structural factors. Values run from -1 to 1; 1 denotes exactly the same photos.

- **Natural Image Quality Evaluator (NIQE)**: By use of statistical feature comparison with a model of natural scene statistics, NIQE is a no-reference metric evaluating image quality. Reduced scores point to better quality.

- **Learned Perceptual Image Patch Similarity (LPIPS)**: LPIPS uses pretrained deep neural network (e.g., VGG) feature-based perceptual similarity evaluation. Lower scores suggest more similarity

- **Fréchet Inception Distance (FID)**: FID evaluates the similarity between distributions of generated and real data using features from a pretrained Inception V3 network. Lower scores indicate closer similarity. The equation is:

$$\text{FID} = \| \mu_r - \mu_g \|_2^2 + \text{Tr}(\Sigma_r + \Sigma_g - 2(\Sigma_r \Sigma_g)^{1/2})$$

where $\mu_r$, $\mu_g$ are mean feature vectors, and $\Sigma_r$, $\Sigma_g$ are covariance matrices of real and gener-ated data features.

### 2.5.2 Qualitative Analysis Techniques

Qualitative analysis played a crucial role in evaluating the performance of the proposed model, particularly in the absence of ground-truth images for real-world blurry images. Human visual assessment was used to assess changes in sharpness, structural consistency, and artifact reduc-tion by comparing the improved outputs with the corresponding blurry inputs. Furthermore, user surveys and questionnaires were done, with participants—both experts and non-experts—asked to rate the perceptual quality of the improved images based on factors such as clarity, naturalness, and visual appeal. These subjective judgments facilitated the evaluation of the model's performance when quantitative comparisons were not possible, providing vital new insights into its application in practical scenarios.

## 2.6 Conclusion

This chapter discusses the background studies, tools required for this research, and existing works relevant to the investigation of facial picture deblurring. It establishes a solid foundation for the thesis by emphasizing key concepts in digital image processing, deep learning, and spe-cialized methods for face image restoration. The literature review highlighted both general and face image deblurring advancements, as well as significant limitations that the proposed work intends to address. Based on the insights gained from this background research, the method-ologies and instruments under discussion provide a solid foundation for the following chapters, which will delve deeper into the proposed method. In the next chapter, the proposed method-ology of this research will be discussed.

# Chapter 3

# Methodology

## 3.1 Introduction

This chapter, which focuses on the restoration of facial photos damaged by blur and noise, provides a detailed description of the methodological approach employed in this work. The primary goal is to create a robust technique for face picture deblurring that makes use of se-mantic information to improve the reconstruction of fine-grained facial traits while preserving identity-specific properties. The chapter aims to provide a complete understanding of the tech-nological foundations of the proposed strategy, clarifying design decisions, justifications, and relationships between the many aspects. It ensures clarity and understanding of the technique's evolution and use by providing a detailed description of each module within the proposed model, the experimental setup, and the reasoning behind their selection.

## 3.2 Dataset Description

In this work, we made use of high-quality facial images with a semantic face mask obtained from the CelebA-Mask-HQ dataset [2]. It featured 30,000 high-resolution face photos of 10,177 distinct celebrities, each accompanied by finely detailed segmentation masks for facial features including eyes, nose, mouth, and hair. Designed for tasks including face parsing, facial attribute analysis, and image synthesis, CelebAMask-HQ offered exact annotations, therefore perfect for training and evaluation of models in computer vision applications needing fine-grained facial feature segmentation. Figure 3.1 shows some sample of the choosen dataset.

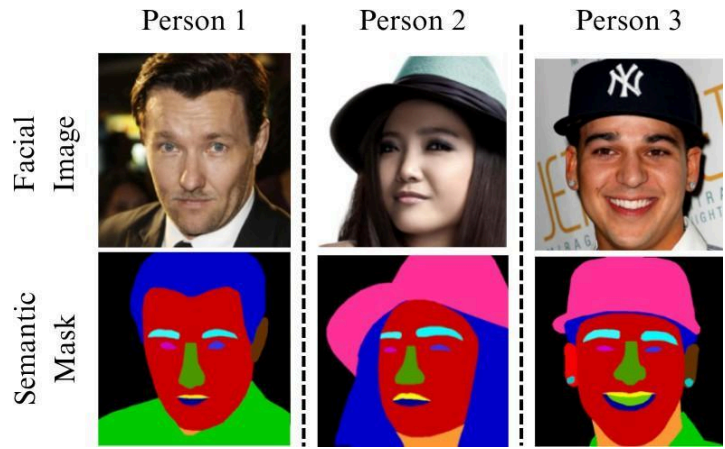

Figure 3.1: Some Sample Images with their Corresponding Masks

## 3.3 Proposed Workflow of Methodology

In this work, we presented a dual input architecture under influence from UNet design. The blurry image and a semantic facial mask are two inputs for this architecture. We developed blurring faces using a randomized blurring algorithm. Semantic face masks depict the position and form of many facial traits including eyes, hair, nose, lips, etc. The suggested model fused the semantic face mask and blurry face at several phases using the multi-level feature fusion technique. Figure 3.2 shows the work flow of our technique.

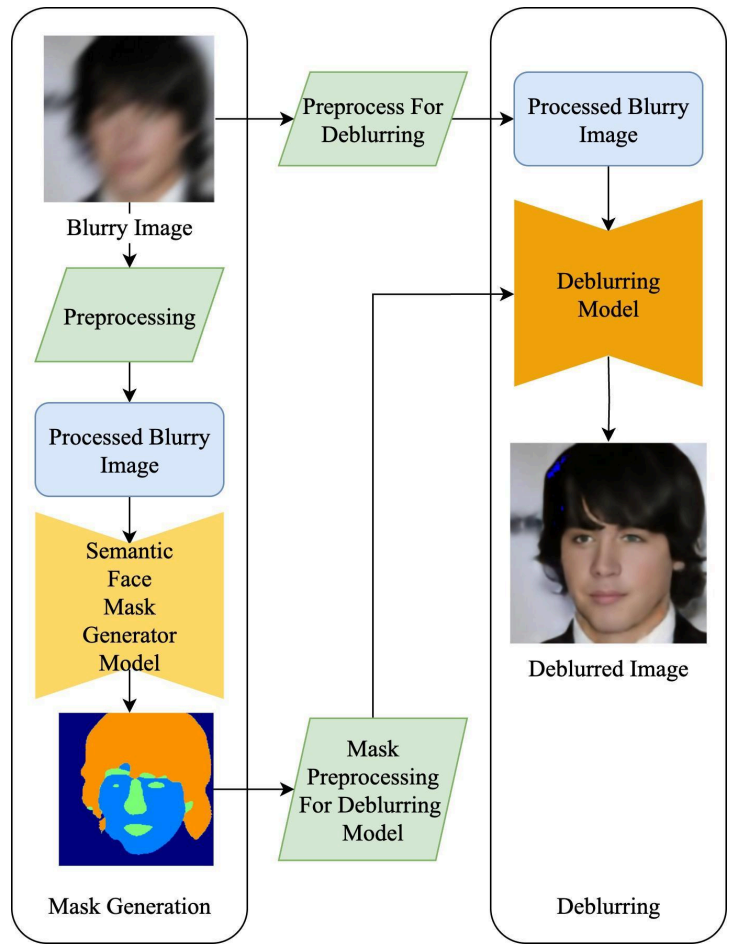

Figure 3.2: Flow of Our Methodology

## 3.4 Proposed Methodology

The proposed methodology began by randomly selecting a subset from a dataset, which is ex-plained in detail in 3.4.6. As a complete set of 30,000 images was not required for the maximum score, it was reduced for computational efficiency. These images were then blurred using the proposed method and resized to 256 x 256 pixels. For mask generation, the original images were converted to grayscale, followed by additional preprocessing steps on the masks. Two models were introduced: one for generating masks from blurred inputs and another for deblur-ring images using the generated masks. Each model employed a slightly different experimental setup. A somewhat different experimental setup was used for training two models.

### 3.4.1 Blurring the Images

There were no blurry photos in the dataset this study used. We thus synthetically generated damaged photos fit for training and evaluation by using blurring methods. In image processing, blurring is a common degradation process usually understood as the convolution of an original image *I* with a blur kernel *k*, resulting a blurred picture *B*. Mathematically, this is written as:

$$B(x, y) = (I * k)(x, y) = \sum_{i,j} I(x - i, y - j) \cdot k(i, j)$$

where ∗ denotes the convolution operation.

In this work, we simulated various and realistic image deterioration situations using a ran-domized blurring technique. Each of the several steps in the process—heavy blur application, resolution degradation, noise addition—introduces notable variation to the produced data.

*Randomized Heavy Blur:* The degradation pipeline began with the application of a ran-domized heavy blur. A random number of blur layers is applied, with the number of layers *n* selected from {1, 2, 3}, providing **3 possible choices** for the number of layers.

For each blur layer, there were **4 possible sequences of operations**: Motion blur only (M), Gaussian blur followed by motion blur (G→M), Motion blur followed by Gaussian blur (M→G)
and Gaussian blur, followed by motion blur, then another Gaussian blur (G→M→G).

Each blur operation involves specific parameter selections: **Gaussian blur** used a kernel size $k_g$ chosen from 6 predefined odd values: {15, 21, 25, 31, 35, 41}. **Motion blur** used a kernel size $k_m$ selected from the same 6 values, combined with 4 possible directions (horizontal,
vertical, diagonal, anti-diagonal), yielding 6 × 4 = 24 combinations.

The total number of combinations for each sequence type was as follows:

- Motion blur only (M): 24

- Gaussian blur followed by motion blur (G→M): 6 × 24 = 144

- Motion blur followed by Gaussian blur (M→G): 24 × 6 = 144

- Gaussian blur, motion blur, then another Gaussian blur (G→M→G): 6 × 24 × 6 = 864

Thus, each blur layer could have a total of:

$$24 + 144 + 144 + 864 = 1,176 \text{ combinations.}$$

Considering the possible number of layers, the total number of heavy blur combinations was:

- For 1 layer: $1,176$

- For 2 layers: $1,176 \times 1,176 = 1,382,976$

- For 3 layers: $1,176^3 = 1,626,943,776$

Summing these gave the total number of heavy blur combinations:

$$1,176 + 1,382,976 + 1,626,943,776 = 1,628,328,728$$

*Resolution Degradation:* After blurring, the image underwent degradation in resolution. It is downscaled by a random scale factor $s$ in the range $[2.0, 4.0]$ and then upscaled back to its original resolution, simulating low-resolution artifacts. Although $s$ is continuous, we discretize it in increments of 0.1 for practical estimation, resulting in:

$$(4.0 - 2.0)/0.1 + 1 = 21 \text{ possible values.}$$

*Gaussian Noise Addition:* Gaussian noise was subsequently added to the image to simulate sensor noise. The noise level $\eta$ was randomly selected from a continuous range between 5.0 and 10.0. For estimation, we discretized this range in increments of 0.1, resulting in:

$$(10.0 - 5.0)/0.1 + 1 = 51 \text{ possible values.}$$

The noisy image $B_{\text{noisy}}$ is given by:

$$B_{\text{noisy}}(x, y) = B(x, y) + \eta \cdot \mathcal{N}(0, 1),$$

where $\mathcal{N}(0, 1)$ is Gaussian noise with zero mean and unit variance.

*Overall Possible Combinations:* The total number of distinct combinations produced by the degradation pipeline could be computed by multiplying the possibilities from each step:

$$\text{Blur combinations} = 1,628,328,728$$

$$\text{Scale factor options} = 21$$

$$\text{Noise level options} = 51$$

Thus, the total number of distinct combinations was:

$$1,628,328,728 \times 21 \times 51 = 1,743,940,018,728 \approx 1.74 \times 10^{12}$$

Discretized values for the scale factor and noise level form the foundation of this approxi-mation. Practically, given these criteria were continuous, there would be theoretically endless combinations. Nevertheless, the discretization offered a decent approximation that emphasizes the great unpredictability brought about by the noise addition, resolution degradation, and ran-domized blur layers. Training effective face image deblurring models able to generalize to real-world degradation depends on this heterogeneity.

### 3.4.2 Pre-Processing

The semantic face mask shows face elements including nose, ear, eye, hair, or any wearings. To reflect all these components, there were nineteen different kinds of pixel values. Simply said, 19 pixel values were merged to only 5 values. A single pixel value allowed one to depict nose, right eye, left eye, right eyebrow, left eyebrow, right ear, left ear, upper lip, and lower lip. One pixel value indicated wearings including a hat/cap, nose pin, or earrings. Another pixel value stood in for the face. Additionally shown was hair using another single-pixel value. Including the backdrop, everything else was set to 0 pixel value. Figures 3.3 illustrate the mask both before and after this grouping. The blurry image, the semantic face mask, and the original photos were then scaled into a 256x256x3 form. On the semantic mask creation, however, the blurry face was changed to grayscale and the semantic mask was one hot encoded. Dividing every pixel by 255 helped to normalize both the original and the blurry images.

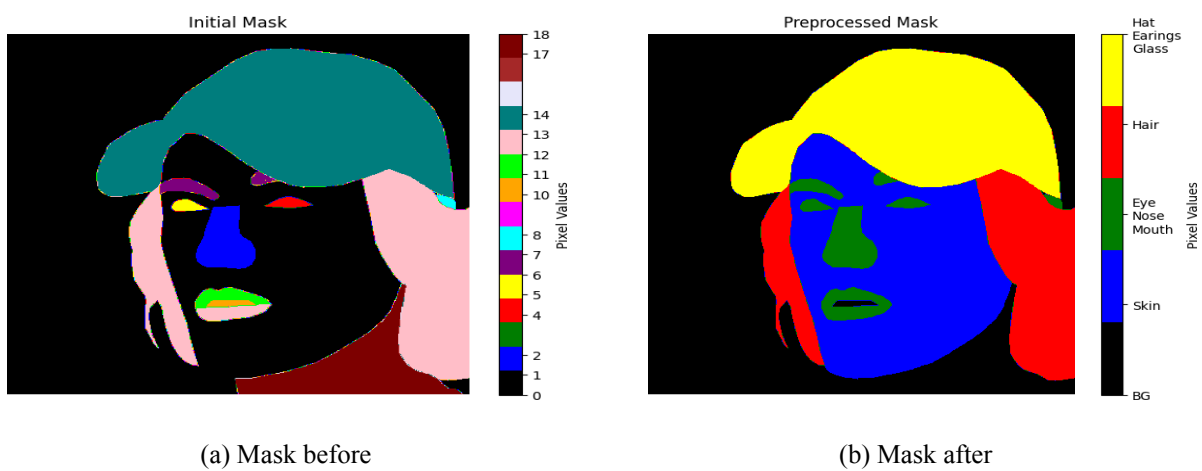

(a) Mask before          (b) Mask after

Figure 3.3: Merging Mask Labels into Five Groups

### 3.4.3 Different Modules Used in Our Proposed Model

We deblurred blurry facial images using two CNN architectures. One was to create the semantic face mask from a blurry image and the one was to create the deblurred image from the blurry image by means of the generated semantic face mask. Since both models were constructed from the same modules, consistency and efficiency are guaranteed. In the next sections, we discuss their design, training, and performance, stressing their success in facial image restoration.

#### 3.4.3.1 Residual Dense Convolution Block (RDC)

The Residual Dense Convolution Block (RDC) was a core component of our proposed archi-tecture, designed to enhance feature extraction and propagation through dense connections and residual learning. The RDB began by iteratively applying convolutional layers, each followed by batch normalization and an activation function (e.g., ReLU). These layers were densely con-nected, ensuring that features from all previous layers are reused, promoting rich feature repre-sentation. For each layer $i$, the output $x_i$ was computed as:

$$x_i = \text{Activation}(\text{BatchNorm}(\text{Conv2D}(x_{i-1}))),$$

where $x_{i-1}$ was the concatenated output of all previous layers. After the dense layers, the concatenated feature maps were passed through a $1 \times 1$ convolutional layer to reduce dimen-sionality and fuse features. This operation can be expressed as:

$$x_{\text{fused}} = \text{Conv2D}_{1 \times 1}(\text{Concat}(x_0, x_1, \ldots, x_n)),$$

where $x_0$ is the original input and $x_1, \ldots, x_n$ were the outputs of the dense layers. To pre-serve low-level features and improve gradient flow, the original input was passed through a $1 \times 1$ convolutional layer and added to the dense output, creating a residual connection. The final output $y$ of the RDB is computed was:

$$y = \text{Activation}(\text{Add}(x_{\text{fused}}, \text{Conv2D}_{1 \times 1}(x_0))).$$

Among numerous benefits, the RDB block provided feature reusing through dense connec-tions, enhanced gradient flow via residual learning, and effective feature fusing utilizing $1 \times 1$ convolution. Extraction and propageneration of hierarchical features depended critically on our dual-input architecture, which combined a semantic face mask for facial image deblurring with a blurry image. Leveraging its capacity to maintain low-level and high-level features, the RDB helped the model to recover fine details and structural information from degraded inputs, thereby producing clear, high-quality images. This made getting strong performance in facial image deblurring tasks dependent on the 6 RDB. A graphical diagram is shown in Figure 3.4.

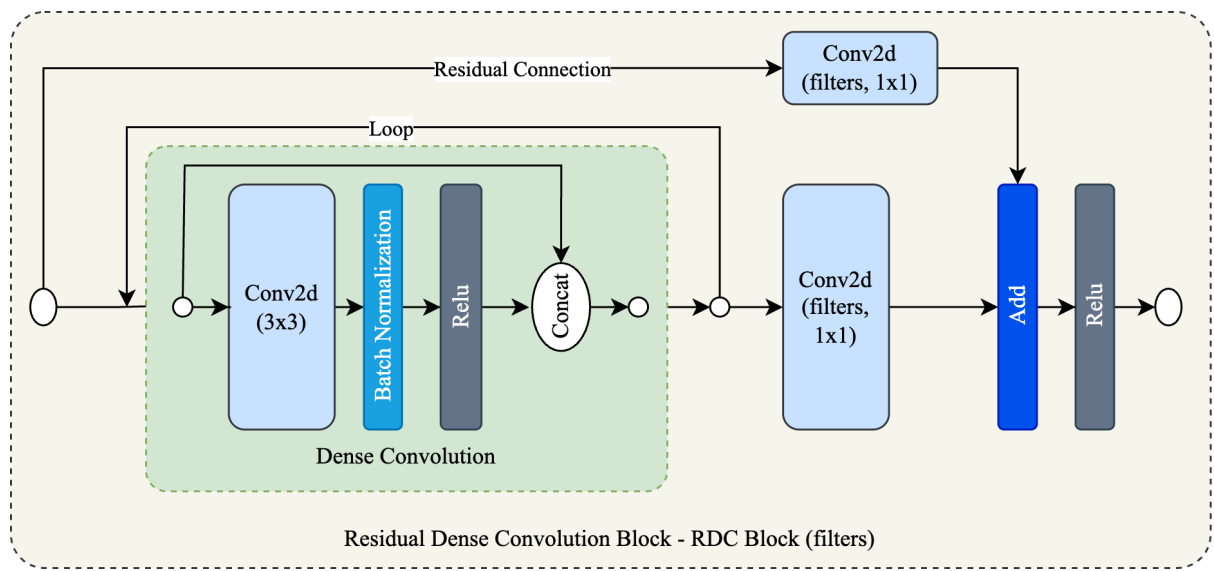

Figure 3.4: The Residual Dense Convolution Block

### 3.4.3.2 PixelShuffle Upsample

Designed to rapidly raise the spatial resolution of feature maps, the **PixelShuffle Upsample** block [28] Starting with a convolutional layer that increases the number of channels by a factor of scale$^2$, where scale is the intended upsampling factor (e.g., 2 for 2x upsampling), it moves through The **PixelShuffle** method then reorganized the enlarged channels into spatial dimen-sions, hence raising the feature map's resolution. This procedure has mathematical expression as:

$$x_{\text{upsampled}} = \text{DepthToSpace}(\text{Conv2D}(x)),$$

where DepthToSpace is the PixelShuffle operation and Conv2D was the convolutional layer. The upsampled feature map was then passed through a ReLU activation to introduce non-linearity. This approach was computationally efficient and reduced artifacts compared to tradi-tional transposed convolutions, making it ideal for tasks requiring high-resolution feature maps.

### 3.4.3.3 Attention Upsample

The **Attention Upsample** block integrated an attention method to improve skip connections, hence augmenting the upsampling process. Following PixelShuffle Upsample to upsampling the input feature map, a **Convolutional Block Attention Module (CBAM)** handles the skip connection, which transports high-resolution features from previous layers. CBAM guaranteed that the skip connection adds significant information by using both channel and spatial attention to emphasize key elements and suppress irrelevant ones. We derive the attention-enhanced skip link was:

$$\text{skip\_attention} = \text{CBAM}(\text{skip\_connection}).$$

The upsampled feature map and the attention-enhanced skip connection were then concate-nated to fuse multi-scale features:

$$x_{\text{fused}} = \text{Concat}(x_{\text{upsampled}}, \text{skip\_attention}).$$

A summarized graphical representation is shown in Figure 3.5.

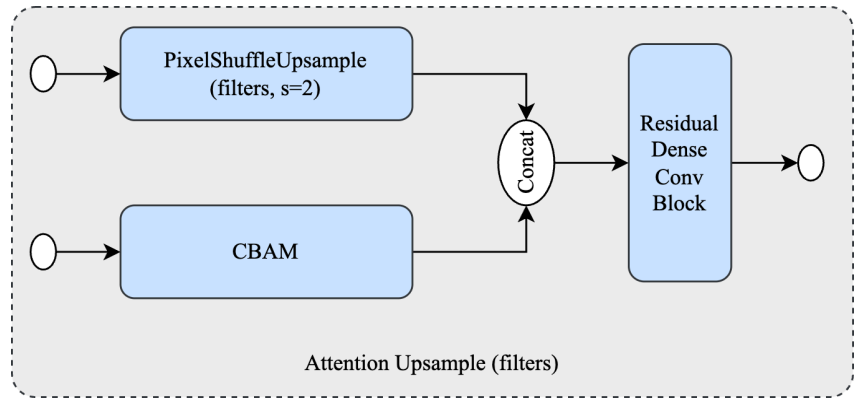

Figure 3.5: The Attention Upsample Block

### 3.4.3.4 Convolutional Block Attention Module (CBAM)

By progressively applying channel and spatial attention, the **Convolutional Block Attention Module (CBAM)** refined feature maps. Channel attention computed weights that highlight significant channels using global average and max pooling, then dense layers using ReLU and sigmoid activations. These weights scaled the supplied feature map. To produce weights, spatial attention collects average and max pooled features, runs them through a convolutional layer with sigmoid activation, then scaled the feature map to concentrate on important spatial locations. For specifics, a detailed diagram is shown in Figure 3.6.

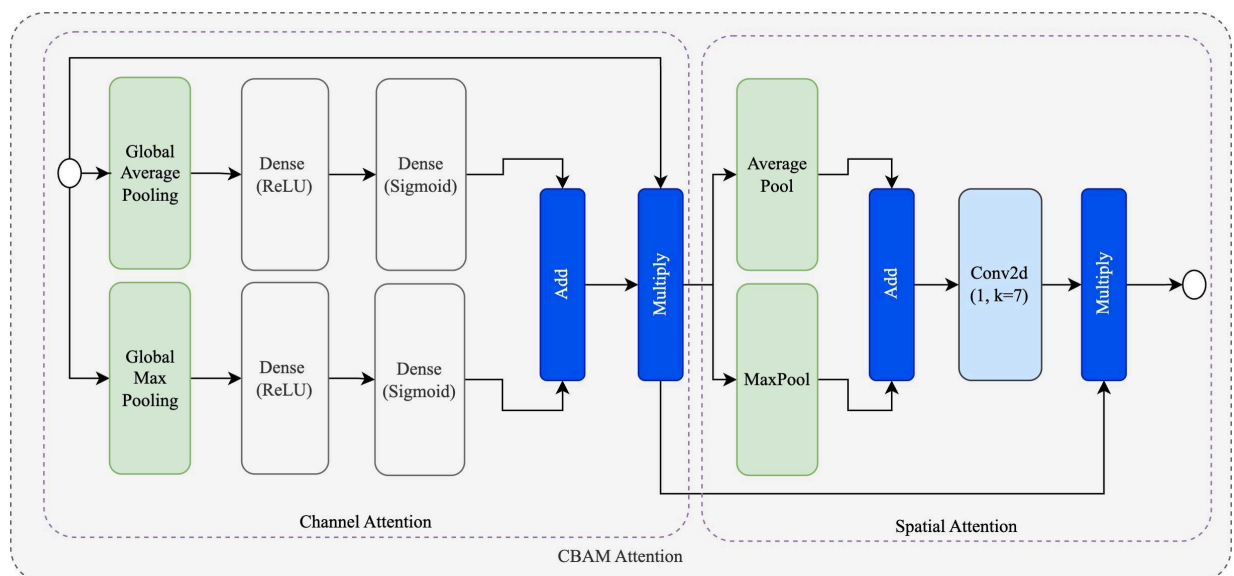

Figure 3.6: The Convolutional Block Attention Module

#### 3.4.3.5 Post-Processing Output

The **post-processing** tool modified output image brightness and contrast. First, one computed the image's mean intensity $\mu$. Scaling the variation from the mean by a factor $c = 2.0$ accen-tuated the contrast; adding an offset $b = 0.1$ increases the brightness. Mathematically, this can be shown as:

$$\text{output} = (\text{output} - \mu) \cdot c + \mu + b.$$

This simple yet effective post-processing step improves the visual quality of the output image.

### 3.4.4 Architecture of Semantic Mask Generator

Developed as a fundamental component to extract a semantic face mask from a blurry input image, the semantic mask generator proves indispensable for the deblurring approach. We used a modified U-Net-like architecture with an encoder-decoder paradigm. Multi-scale features were obtained using an image branch in which hierarchical representations were generated by means of four convolutional layers. The decoder then handled these representations such that spatial information was restored. Features from related encoder layers were merged by stacking one over another on the channel axis at each level using an attention-guided upsampling tech-nique, then Residual Dense Convolution Blocks arranged with channel sizes of 256, 128, and 64, respectively. This design conserved structural integrity across sizes and improved feature refinement.

The decoder output was then processed using a final convolutional block including an at-tention module and a Residual Dense Convolution block with 32 channels to give important facial features top priority. Following an extra 1x1 convolution to create the refined semantic mask, a 1x1 convolution with softmax activation produced a segmentation mask fit for 19 facial classes. This was thus set up as a stand-alone model, with the semantic mask output essentially defining facial components and a blurry image received as input. The final mask guaranteed the effectiveness of the framework in reconstructing high-quality facial images, therefore offering critical direction for the deblurring stage. Figure 3.7 shows a thorough graphical overview.

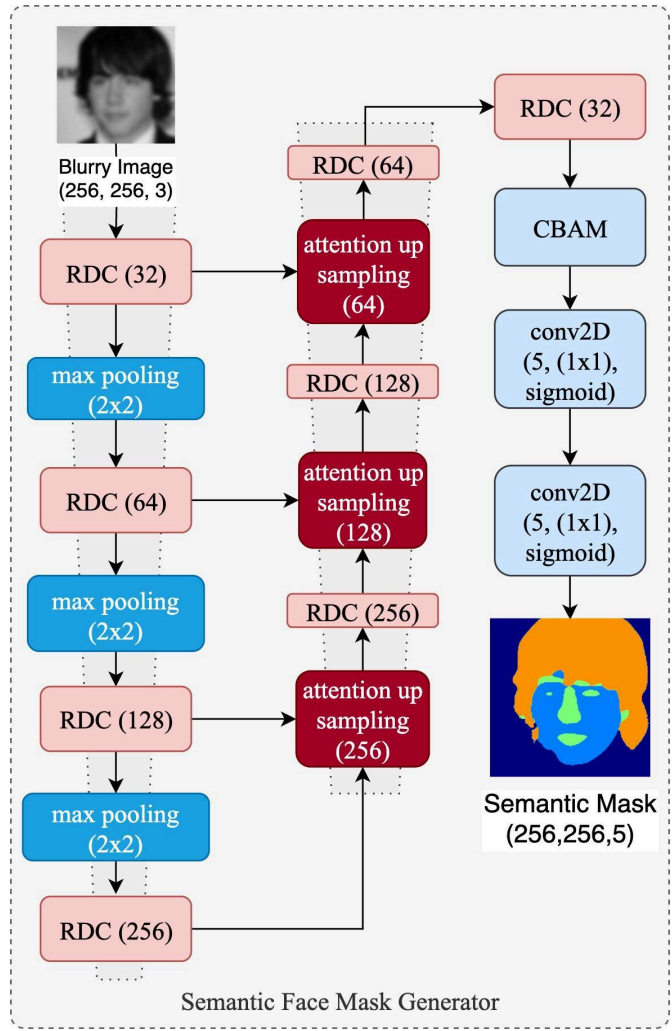

Figure 3.7: Semantic Mask Generator Model Architecture

The deblurring UNet was built lighter than traditional UNet. Table 3.1 shows the parameter counts of this model.

Table 3.1: Mask Generator Model Parameter Summary

| Parameter Type | Count (Size) |
| --- | --- |
| Total Parameters | 5,416,159 (20.66 MB) |
| Trainable Parameters | 5,414,751 (20.66 MB) |
| Non-trainable Parameters | 1,408 (5.50 KB) |

### 3.4.5 Architecture of Deblurring Network

In this work, one of our major contributions was the introduction of SMFD-UNet (Semantic Mask Fusion Deblurring UNet), a lightweight architecture designed to deblur a blurry and noisy facial image. This model introduced a revolutionary way to take assistance of the semantic face mask to deblur the blurry and noisy image. The model architecture is shown in Figure 3.8

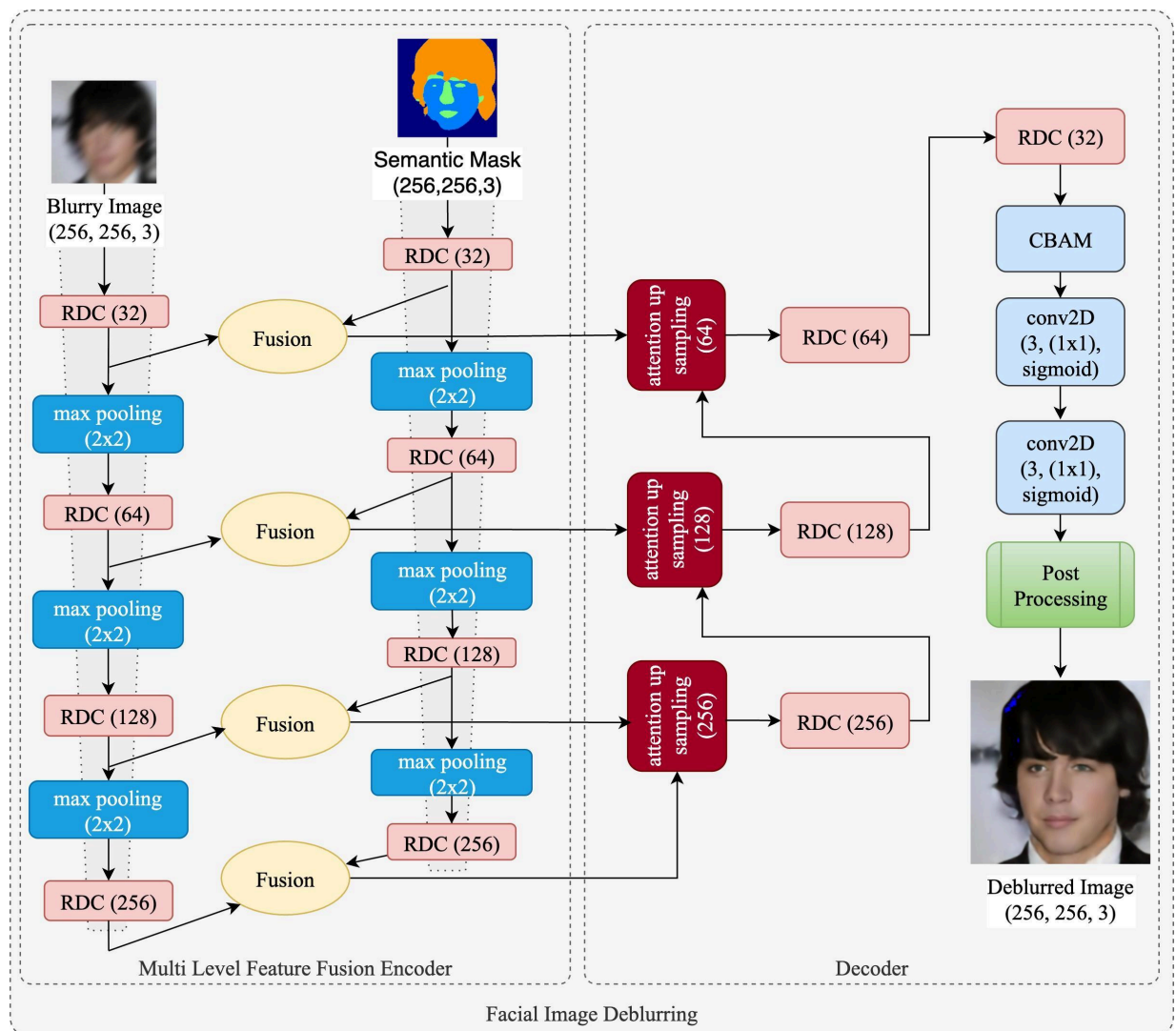

Figure 3.8: SMFD Model Architecture

The SMFD-UNet was also built very lighter compared to existing deblurring models. Table 3.2 shows the parameter counts of this model.

Table 3.2: SMFD-UNet Parameter Summary

| Parameter Type | Count (Size) |
| --- | --- |
| Total Parameters | 7,532,601 (28.73 MB) |
| Trainable Parameters | 7,531,641 (28.73 MB) |
| Non-trainable Parameters | 960 (3.75 KB) |

#### 3.4.5.1 Encoder

The model consisted of two encoder branches: one for processing the semantic face mask input and another for the input image. With feature extraction using **Residual Dense Convolution Blocks (RDC)** both branches had a similar architecture. Every encoder branch consisted of four convolution and pooling steps. The input is first run through an RDB to extract hierarchical features, then max-pooling helps to shrink spatial dimensions at every level. Concatenated by stacking one over another on the channel axis, outputs of matching stages from both branches were fused at different stages, therefore allowing the model to efficiently employ both visual and semantic aspects.

#### 3.4.5.2 Decoder

From the fused feature maps, the decoder rebuilded the high-resolution output. It uses **Attention Upsample** blocks to combine upsampling via the PixelShuffle technique with attention meth-ods to hone skip connections. Using **Convolutional Block Attention Module (CBAM)**, the upsampled features were combined with skip connections from the encoder so stressing signif-icant features. An RDB followed every upsampling stage to improve the features even further. This progressive upsampling and refining technique guaranteed the structural information's and fine detail's recovery.

#### 3.4.5.3 Final Layers

A sequence of convolutions and post-processing techniques made up the latter levels of the model. To improve feature representations, the decoder's output was passed via a CBAM layer and a last RDB. The segmentation result is generated by then using a 1×1 convolution. Another 1 × 1 convolution helped to improve this output; next, a post-processing phase modified the resulting image's brightness and contrast. The last result was a polished, high-quality image fit for jobs such segmentation or deblurring of facial images.

### 3.4.6 Experimental Setup

We proposed two architectures to deblur facial images holistically in this work. A blurry facial image is used to construct a semantic face mask in the first architecture. The second archi-tecture, the Semantic Mask-Guided face Deblurring (SMFD) model, creates a

high-quality de-

blurred face image from the blurry image and the semantic mask. We trained our main de-blurring model, SMFD, using dataset ground-truth semantic face masks and our semantic mask generating model masks to objectively evaluate their efficacy. This dual-training strategy helps us assess the SMFD model's semantic input quality reliability and performance. Though the SMFD deblurring model was trained on 2,000 samples, we trained the semantic mask generat-ing model on 3,000 samples. Despite differences in assessment metrics and loss functions for semantic mask prediction and picture deblurring, both models' experimental setups are similar. These distinctions ensure that each model is examined according to its goals, optimizing each model and assessing our suggested structure.

#### 3.4.6.1 Real-time Data Augmentation

Data augmentation was necessary to generalize the model. It also helped to reduce overfitting. Our real-time data augmentation included geometric transformations such as random rotations (up to $\pm 30°$), random horizontal flips, and random crops (0.8x to 1.0x). It also included a few photometric augmentations, such as brightness adjustments with a factor of up to 0.7 to 1.3, contrast variations up to a factor of 0.7 to 1.3.

#### 3.4.6.2 Cross-Validation

Five-fold cross-validation verified our model's reliability. The dataset was initially split 80:20 into training/validation and testing sets to properly split each data type. Five equal-sized folds comprised the training/validation dataset. Every cycle, one fold was validated and the model was trained using the other four. Five repeats of this stage ensured each fold was validated once. Each fold received 50 epochs of training in a 16-fold batch. Our training technique re-duced overfitting by using many data sources. The model's efficacy was assessed by combining performance assessments from each fold.

#### 3.4.6.3 Hyperparameter Tuning

A little bit different hyperparameters were used for the Mask Generator and the deblurring SMFD-UNet model. These hyperparameters were set by being influenced by existing works on this task and through a trial-and-error mechanism. We adjusted several hyperparameters to reach optimal performance and training economy. The learning rate was first set to 0.001, same for both models, and a method of decrease was followed. The validation Dice

coefficient

for Mask Generator and validation SSIM for SMFD-UNet were constant for five consecutive epochs, the learning rate was reduced to 20% of its current value. This approach continued until the learning rate dropped to a minimum of $1 \times 10^{-9}$. An early stopping condition was developed to prevent overfitting; it halted training when the validation Dice coefficient for Mask Generator and validation SSIM for SMFD-UNet did not improve for ten consecutive epochs. Moreover, model checkpointing was used during training to save the weights of the model when the validation Dice coefficient for Mask Generator and the validation SSIM for SMFD-UNet increase, thereby preserving the optimal model configuration for the last assessment.

#### 3.4.6.4 Evaluation Metrics and Loss Function

Different evaluation metrics and loss functions were used for the two models. These evaluation metrics and loss functions were described in detail in chapter 2.

**Semantic Mask Generator Model's Evaluation Metrics and Loss Function:** We eval-uated mask generator model using the Dice Coefficient and Jaccard Index. The loss function consisted in dice loss. Every metric was computed channel wise independently before a mean value was derived.

**De-blurring SMFD Model's Evaluation Metrics and Loss Function:** As assessment measures we applied PSNR, SSIM, LPIPS, MSE, NIQE, FID. PSNR assesses output pixel-wise accuracy while SSIM gauges output structural similarity with reference to the genuine value. We employed mean square error as our loss function for this work.

## 3.5 Conclusion

Using the CelebA dataset with synthetically generated blur through a randomized pipeline pro-ducing 1.74 trillion permutations, this chapter described the proposed technique for face picture deblurring. Combining blurry pictures with semantic face masks, preprocessed to 256x256x3 with simplified 5-value masks, the proposed SMFD-UNet is a dual-input U-Net-based model. Important elements include Residual Dense Convolution Blocks, PixelShuffle Upsample, At-tention Upsample, and CBAM for efficient feature extraction and refining; then, post-processing improves visual quality. This method addresses identity retention and texture shortage, there-fore laying a good basis for experimental analysis. Performance evaluation of this technique will be covered in the future chapter.

# Chapter 4

# Result Analysis and Discussion

## 4.1 Introduction

This chapter provides a comprehensive evaluation of the proposed Semantic Mask Fusion De-blurring UNet model, which reconstructs facial images with semantic face masks generated directly from image blurriness. The performance was evaluated using peak signal-to-noise ra-tio (PSNR) and structural similarity index (SSIM), as well as qualitative visual comparisons to state-of-the-art algorithms such as CodeFormer, RestoreFormer, RestoreFormer++, and GFP-GAN. The examination also includes an ablation study to emphasize the contributions of key components like the semantic mask branch and attention-based upsampling. Training valida-tion curves and real-world deblurring results were also examined in order to demonstrate model generalization and strength. The results demonstrate how successfully the lightweight design provides remarkable structural similarity and competitive naturalism, overcoming the concerns of face image deblurring raised in previous chapters.

## 4.2 Results Analysis

### 4.2.1 Result Obtained by Our Model

Our approach mostly depended on the acquisition of ideal semantic face mask. Still, we sug-gested a technique to create face masks from the blurry pictures. Still, the replica differs from the original in some ways. To deblur facial photos, we thus have compared our deblurring technique with produced masks and original masks provided in the dataset.

#### 4.2.1.1 Result in Generating Semantic Face Mask

Our semantic mask generation was doing pretty good. It achieved satisfactory dice co-efficeint of 64.84% and jaccard index of 55.72%. The sores were not that much good. This was due to the severe blurriness that was causing misinterpretation of a few portions. A visual representation of our score is given in Figure 4.1.

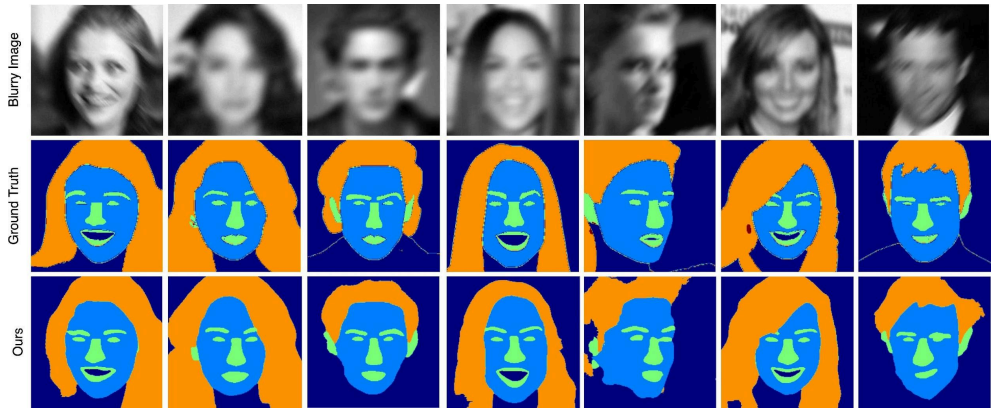

Figure 4.1: A Visual Representation of Our Result in Generating Semantic Face Masks

A comparative study was done with two state-of-the-art models for segmentation, UNet and UNet++ [54]. Table 4.1 shows the performance of each model along with its parameters. This table shows that our model, although it achieved lower scores, is computationally more efficient, which aligned more closely with our objective of making the architecture lightweight. Also, the differences in scores were negligible.

Table 4.1: Quantitative Comparison of Our Model with Recent State-of-the-art Segmentation Models

| Model | Dice ↑ | Jaccard ↑ | Parameters ↓ |
| --- | --- | --- | --- |
| UNet | 67.21% | 58.00% | 31,401,605 |
| UNet++ | **68.20%** | **58.74%** | 36,643,333 |
| **Ours** | 64.84% | 55.72% | **5,416,159** |

#### 4.2.1.2 Result in Deblurring Facial Images

Using Peak Signal-to- Noise Ratio (PSNR) and Structural Similarity Index Measure (SSIM), the performance of the proposed Semantic Mask Fusion Deblurring UNet was assessed against many state-of- the-art techniques including CodeFormer, RestoreFormer, RestoreFormer++, and GFPGAN. The findings, collected in Table 4.2, show how well our model restores noisy and blurry facial images guided by semantic face masks and multi-stage feature fusion.

Table 4.2: Quantitative Comparison of Our Model with Recent State-of-the-art Deblurring and Restoration Models by Zero-shot Prediction

| Model *(Publisher, Year)* | PSNR ↑ | SSIM ↑ | MSE ↓ | NIQE ↓ | LPIPS ↓ | FID ↓ |
|---|---|---|---|---|---|---|
| CodeFormer *(NeurIPS, 2022)* | 22.59 | 61.80% | 417.59 | **0.2565** | 0.3211 | 119.31 |
| RestoreFormer *(CVPR, 2022)* | 22.91 | 63.28% | 388.53 | 0.2613 | 0.3450 | 129.25 |
| RestoreFormer++ *(TPAMI, 2023)* | 22.88 | 62.65% | 395.52 | 0.2599 | 0.2368 | **89.70** |
| GFPGAN *(CVPR, 2021)* | 22.99 | 63.70% | 381.64 | 0.2644 | **0.2295** | 97.42 |
| **Ours with Original Mask** | **24.37** | **70.81%** | **330.19** | 0.2664 | 0.2933 | 110.45 |
| **Ours with Generated Masks** | **23.82** | **69.34%** | 390.07 | 0.2682 | 0.3251 | 123.54 |

As we suggested a dual input model for deblur facial photographs. Among its inputs is semantic facial mask. We therefore also suggested a semantic face mask generating model from blurry images. We also examined result using original face mask to ignore the dependence with mask generator model, so verifying the deblurring model how good it is.

Figure 4.2 shows a visual comparison of our model with the state-of- the modern mod-els. These findings show the benefit of including semantic face mask guidance with dual-input multi-stage feature fusion in order to enable more exact reconstruction of facial details and enhanced perceptual quality.

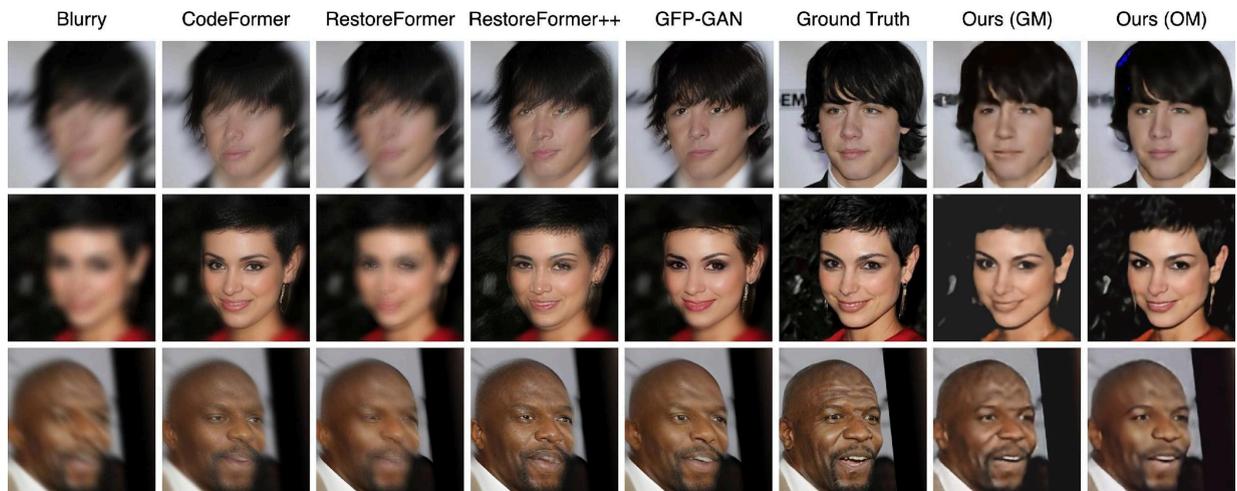

Figure 4.2: A Visual Comparison of Our Result with Other Recent State-of-the-art Models by Their Zero-shot Prediction

A few real-world blurry photographs, images containing motion blur mainly, were taken of a famous actor, Rowan Atkinson, from the sets of a Hollywood movie named "Johnny En-glish" [55]. The results of deblurring these photographs using our suggested model and other state-of-the-art models are provided in Figure 4.3. These samples illustrate that our models' deblurred photos were not so clear, but it has improved the structural likeness of the human. Other state-of-the-art models deblurred photos are sharp, but they are synthetic images where structural dissimilarity may be found, and there is also a change of facial expression. As for these real-world blurry images, no ground truth was present, hence no quantitative compari-son was possible. So, a qualitative evaluation was performed by taking a survey among a few people, and their votes also match the above explanations.

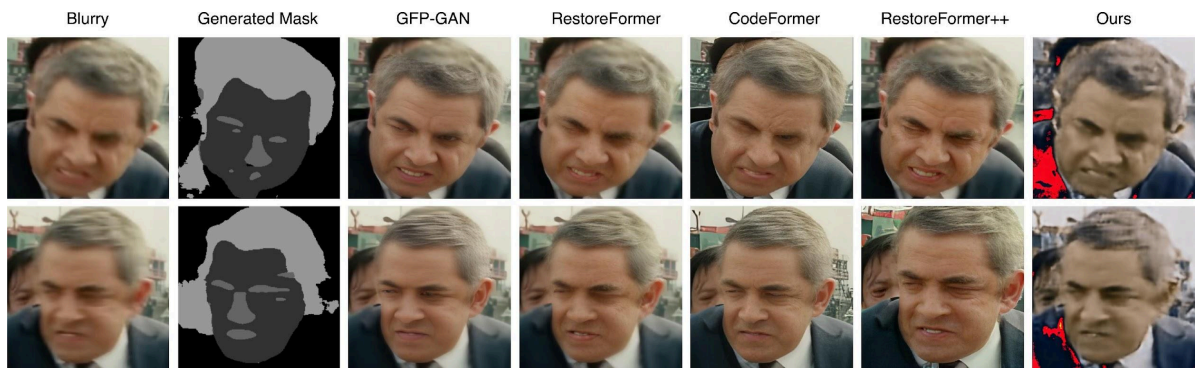

Figure 4.3: Deblurring Real-world Blurry Images

The enhancements in PSNR and SSIM emphasize the model's ability to effectively reduce noise and recover fine facial features, proving the efficacy of the suggested architecture in ad-dressing the issues of facial deblurring with a lighter architecture. Lighter because existing works like GFPGAN, RestoreFormer, RestoreFormer++, CodeFormer, and others are mainly GAN, Diffusion, or Transformers, and they are typically heavier than CNNs and need a huge collection of data to be well trained, and also a huge amount of time. Additionally, our proposed model was trained well with a smaller dataset, and performance across real-world blurriness proved in favor of this.

In Figure 4.4, the training and validation curves of the training time of the deblurring SMFD-UNet model are shown. From the curves, it is clear that the model is well-trained. Since there are no significant gaps between the training and validation curves, the SSIM validation curve is superior to the training curve, supporting the model's generalization capability. Addition-ally, both training and validation scores are satisfactory, indicating no issues with overfitting or underfitting. Additionally, the smooth, incremental curves reflect the optimal selection of hyperparameters.

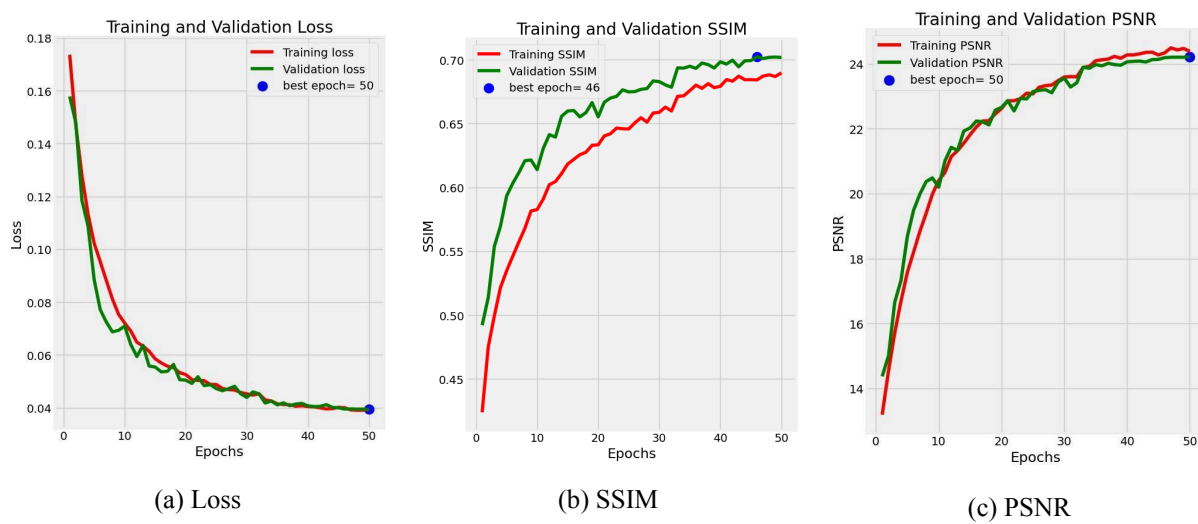

(a) Loss  (b) SSIM  (c) PSNR

Figure 4.4: Training and Validation Curves

### 4.2.2 Ablation Study

An ablation study was performed by methodically changing the design and monitoring the per-formance using Peak Signal-to-Noise Ratio (PSNR) and Structural Similarity Index Measures (SSIM) in order to assess the influence of every component in our suggested model. Table 4.3 shows the outcomes. Comprising the backbone devoid of the semantic mask branch, the baseline configuration obtained an SSIM of 66.25% and a PSNR of 22.46. By including the semantic mask branch with conventional upsampling, performance was much enhanced, and a PSNR of 23.91 and an SSIM of 69.22% resulted. This shows how much semantic direction helps to improve reconstruction quality.

Table 4.3: Performance Comparison of Different Configurations Based on PSNR and SSIM Metrics

| Configuration | PSNR ↑ | SSIM ↑ |
| --- | --- | --- |
| Backbone without Semantic Mask Branch | 22.46 | 66.25% |
| Backbone + Semantic Mask Branch + Traditional Upsampling | 23.91 | 69.22% |
| Backbone + Semantic Mask Branch + Attention Upsampling (Transpose) | 23.89 | 70.20% |
| Backbone + Semantic Mask Branch + Attention Upsampling (Pixel Shuffle) | 24.14 | 70.42% |
| Backbone + Semantic Mask Branch + Attention Upsampling + Post Processing Block | 24.37 | 70.81% |

Replacing conventional upsampling with attention-based upsampling also showed further results. While pixel shuffle-based attention upsampling somewhat improved performance to a PSNR of 24.14 and an SSIM of 70.42%, transpose-based attention upsampling produced a PSNR of 23.89 and an SSIM of 70.20%. These results emphasize how well attention processes capture fine-grained features during upsampling.

Adding a post-processing block to the attention upsampling (pixel shuffle) setup achieved optimal performance (PSNR 24.37, SSIM 70.81%). The ablation study shows that the se-mantic mask branch, attention-based upsampling, and post-processing block each significantly contribute to the model's performance, with their combined effect yielding the best results.

A layer-wise feature map visualization has been shown for the Mask Generator model in Figure 4.5 and for SMFD-UNet in Figure 4.6. As feature maps could be multi-channel, proper understanding is not possible, though it gives a little understanding of how the proposed archi-tectures reach their destination. For channels like 3, the output is plotted as an RGB image; for 5 channels, anti-one-hot encoding reduces it to a single channel; otherwise, only the first channel is displayed. The outputs in both figures changed over time because of the architecture's de-sign, which uses Residual Dense Connection (RDC) layers to keep and improve features across layers, which helped keep more detail. Max pooling kept important information while reducing spatial dimensions, and attention up-sampling made relevant features stronger, which helped the network produce useful outputs like semantic masks or images that have been deblurred. The model learned more complex patterns as the channel depth slowly increased (for example, from 64 to 256). The fusion stages in the deblurring process combined semantic guidance with image data, which made the images clearer. The network's ability to learn hierarchical features can be seen by the consistent improvement in output quality, which went from blurry to sharp or masked to accurate.

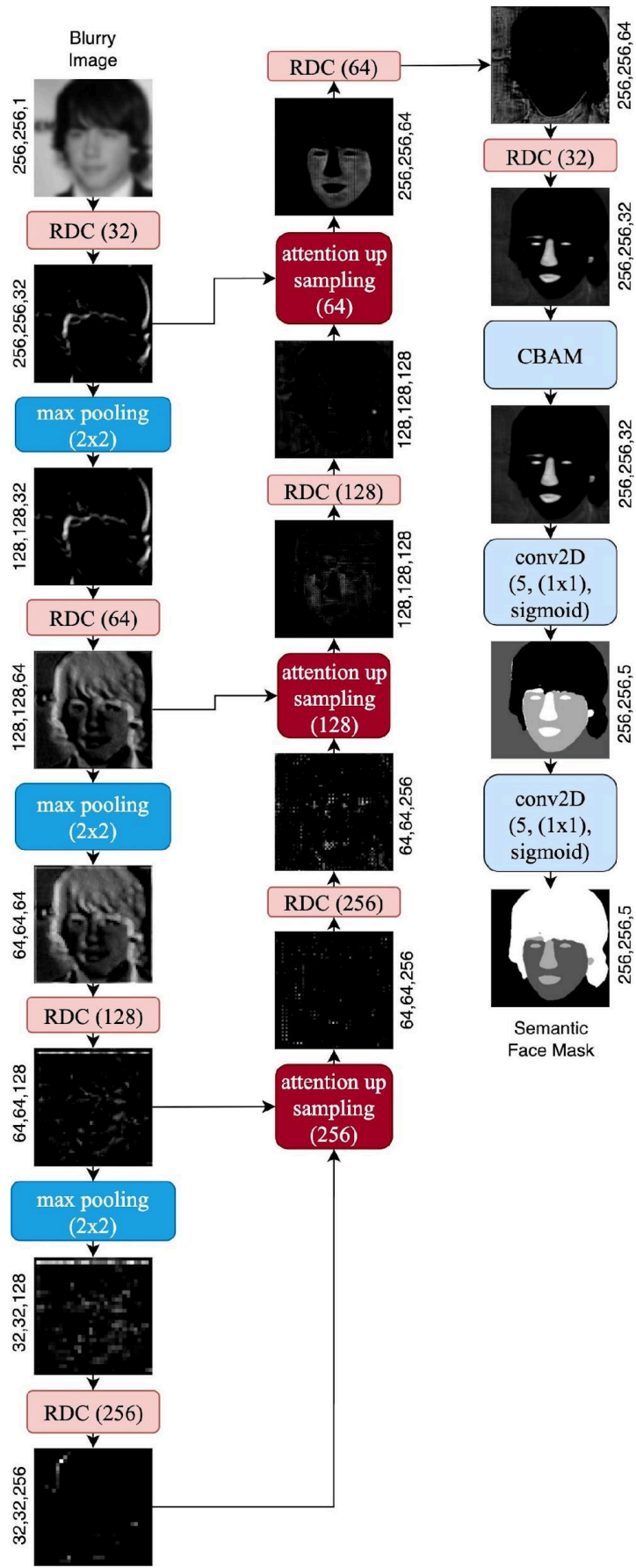

Figure 4.5: Layer-wise Feature Map Visualization for the Mask Generator Model

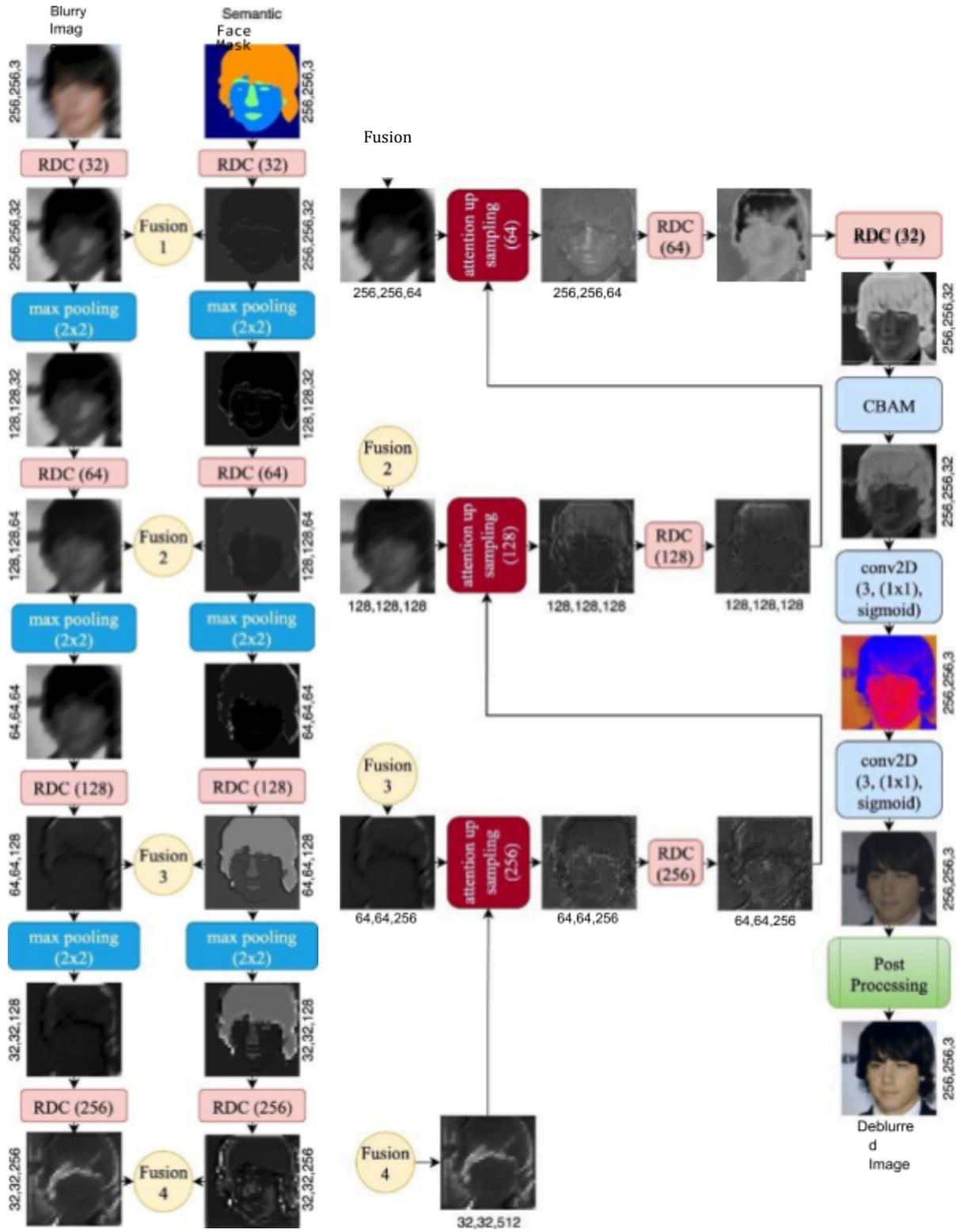

Figure 4.6: Layer-wise Feature Map Visualization for the SMFD-UNet

### 4.2.3 Cross-Validation Scores

At the end, to evaluate the performance of the proposed model, we conducted a 5-fold cross-validation experiment, shown in Table 4.4. Two key metrics were used to assess the quality of the reconstructed images: Peak Signal-to-Noise Ratio (PSNR) and Structural Similarity Index Measure (SSIM). This proves the generalization of our model across diverse kinds of blurry inputs.

Table 4.4: PSNR and SSIM values across folds with mean and standard deviation

| Fold | PSNR | SSIM (%) |
| --- | --- | --- |
| Fold 1 | 24.4046 | 70.61 |
| Fold 2 | 24.1845 | 69.50 |
| Fold 3 | 24.7568 | 71.22 |
| Fold 4 | 24.2840 | 70.27 |
| Fold 5 | 24.7137 | 71.05 |
| **Mean ± SD** | **24.4687 ± 0.2289** | **70.53 ± 0.61** |

## 4.3 Discussion

Our model outperformed all other state-of-the-art restoration models in terms of SSIM and PSNR. Our model's mean square error was superior for deblurring with the original face mask, but it lagged behind the MSE of GFPGANs and RestoreFormer when using the generated face mask. Our model lags behind for every other metric. This is because our model concentrated on structural similarity, while other models prioritized naturalness and sharpness. Other mod-els likely achieved better naturalness due to their emphasis on generative techniques, such as those in GFPGAN and RestoreFormer, which prioritize texture synthesis and sharpness through adversarial training or advanced loss functions like perceptual loss. These approaches enhance visual appeal by generating finer details, often at the cost of structural fidelity, as they may overfit to statistical patterns rather than preserving original structural elements like identity traits and expressions. On the other hand, identity traits and expressions were preserved in our model's deblurred images, despite the fact that they were less crisp and natural. First, the ro-bust multi-stage fusion technique, which combined the semantic masks with the blurry input in the encoder to combine features at each stage, was the reason for our model's performance.

The RDC block then continued to have a significant influence on both the mask generation and deblurring processes. It is made up of residual connections and dense convolutions. Dense convolution enabled feature reuse and better information flow by feed-forward connecting each layer to every one after it. Dense convolution concatenated all previous layer outputs as input, improving gradient flow and minimizing vanishing gradient problems in contrast to traditional convolution, which only receives input from the previous layer. As a result, deeper networks with fewer parameters were made possible by more effective parameter usage. By addressing vanishing gradient problems and enhancing feature propagation, residual connections made it simpler to train deep networks by adding a layer's input straight to its output. Better upsampling with fewer parameters was made possible by pixel-shuffle upsampling. In order to improve fea-ture quality, CBAM used channel and spatial attention to highlight significant high-resolution features and suppress those that weren't relevant. This CBAM feature improved our model's upsampling process when combined with Pixel-shuffle upsampling. The final post-processing block enhanced the perceived sharpness and visual quality.

## 4.4 Conclusion

Validated in this chapter, the Semantic Mask Fusion Deblurring UNet model outperformed state-of-the-art methods, including CodeFormer and GFPGAN, with a PSNR of 24.4687 ± 0.2289 and SSIM of 70.53 ± 0.61 % using natural masks. A Dice coefficient of 64.84% and Jaccard index of 55.72% were obtained by the semantic mask generating model. The ablation in-vestigation underlined the relevance of the post-processing block, attention-based upsampling, and semantic mask branch. These results show how successfully the model repairs blurry facial images with competitive naturalness and enhanced structural similarity. This thesis book will draw to a close in the next chapter.

# Chapter 5

# Conclusion and Future Scopes

## 5.1 Introduction

The main results and applications of the Semantic Mask Fusion Deblurring UNet (SMFD-UNet) model for face image deblurring are compiled in this chapter. It presents a thorough overview of the thesis results, emphasizes the need of the suggested approach, and lists possible future topics of research. The chapter ends by considering the whole influence of the work in advancing facial image restoration and its uses in computer vision.

## 5.2 Thesis Conclusion and Summarization

Using semantic face masks to guide the restoration process, this thesis presented SMFD-UNet, a novel lightweight framework meant to solve important problems in facial image deblurring. The work effectively addressed constraints in current literature, including dependence on high-quality reference images, computational inefficiencies, and the failure to preserve identity under strong blur circumstances. The main results and contributions of this effort are compiled as follows:

- **Novel Deblurring Framework**: Using original masks, SMFD-UNet achieved excep-tional performance with a Peak Signal-to-Noise Ratio (PSNR) of $24.4687 \pm 0.2289$ and Structural Similarity Index Measure (SSIM) of $70.53 \pm 0.61$ % by combining blurry im-ages with semantic face masks under a dual-input architecture, surpassing existing works.

- **Reference-Free Semantic Mask Generation**: A UNet-based semantic mask genera-

tor was built to directly extract facial component masks (e.g., eyes, nose, mouth) straight from blurry images, obtaining a Dice coefficient of 64.84% and a Jaccard index of 55.72%. This solved a main restriction in previous semantic-guided methods by removing the re-quirement for clear reference images.

- **Robust Blurring Pipeline**: By use of combinations of Gaussian blur, motion blur, reso-lution degradation, and noise addition, a randomized blurring pipeline was presented to simulate almost 1.74 trillion deterioration scenarios. This guaranteed the resilience of the model to certain real-world blur circumstances.

- **Lightweight and Efficient Architecture**: The model is light and efficient, compared to existing state-of-the-art models. The Mask Generating model has a parameter count of 5,416,159, and the SMFD-UNet has a parameter count of 7,532,601. Residual Dense Convolution Blocks (RDC), PixelShuffle Upsampling, and Convolutional Block Atten-tion Modules (CBAM) produce a computationally efficient model fit for low-resource environments.

- **Multi-Stage Feature Fusion**: To improve the reconstruction of fine facial characteristics and maintain structural integrity, a unique multi-stage fusion technique was presented to merge semantic and visual aspects at several scales.

The ablation investigation verified the important part each component— semantic mask branch, attention-based upsampling, and post-processing block—plays in obtaining high-quality deblurring results. Tests using real-world blurry photos proved the generalizability of the pro-posed model in real-world settings. SMFD-UNet excels in pixel-wise and structural similarity yet unlike rival models that generate crisper but synthetic outputs with changed facial expres-sions retains competitive naturalness measures (NIQE, LPIPS, FID).

This work offered a strong, reference-free, and effective approach, therefore advancing the field of face image deblurring. Through its lightweight architecture and great performance, its applications encompass facial recognition, forensic analysis, photographic improvement, and medical imaging, thus providing both academic and industry value.

## 5.3 Evaluation of Hypotheses

This study was guided by four hypotheses concerning the efficacy of a lightweight, reference-free model for deblurring heavily distorted facial images using semantic masks. The experi-mental results are evaluated below to assess the extent to which each hypothesis was supported. **H1: Semantic masks generated from blurry images can effectively guide identity-preserving deblurring.** The results moderately support this hypothesis. Semantic masks, derived directly from blurry input images, moderately guided the deblurring process by pre-serving critical facial features such as eyes, nose, and mouth contours. The Dice and Jaccard scores were not satisfactory; that's why it is called moderate.

**H2: Artificial uniform blurriness can mimic real-world nonuniform blurriness to a satisfactory level.** This hypothesis was partially supported. As was tested on real-world blurry images, the outputs were good but not perfect, and obviously retrieval of the original image is never possible.

**H3: Reference-free semantic deblurring performs competitively with reference-based methods.** The findings largely support this hypothesis. The proposed lightweight CNN ar-chitecture was trained and tested using both original semantic masks and generated semantic masks derived from blurry input images, with both configurations yielding closely comparable performance.

**H4: Lightweight CNN architectures can achieve high deblurring quality with lower computational costs.** This hypothesis was fully supported by the experimental results. The proposed lightweight CNN architecture achieved superior deblurring performance compared to state-of-the-art transformer-based and GAN-based models, as evidenced by quantitative metrics and qualitative evaluations.

## 5.4 Future Scopes

The results of this work provide several interesting research paths and provide means for several improvements to further deblurring of facial images and their uses:

- **Keep the Same Color as Input**: Proposed model's output was struggling to keep the same color as the input image. Future works can focus on this.

- **Unwanted Colourful Pixels in Background**: Proposed model sometimes failed to cor-rectly output the background and created these unwanted colorful pixels.

- **Enhanced Semantic Mask Generation**: The semantic mask generator, with a Dice co-efficient of 64.84%, could be improved.

- **Expanded Degradation Models**: The current blurring pipeline covers a vast range of scenarios, but future work could incorporate additional degradation types, such as occlu-sions, complex lighting variations, or atmospheric distortions, to further enhance model generalization.

- **Larger and More Diverse Real-World Datasets**: Training and testing on larger, more diverse real-world datasets could improve the model's generalization, particularly for underrepresented facial features or complex real-world scenarios.

These directions aim to build on the strengths of SMFD-UNet, addressing remaining chal-lenges and expanding its applicability across diverse domains and use cases.

## 5.5   Conclusion

This chapter gave a thorough overview of the contributions of the SMFD-UNet model to fa-cial image deblurring, stressing its better performance in PSNR (24.4687 ± 0.2289) and SSIM (70.53 ± 0.61 %), reference-free semantic mask generation, and strong blurring pipeline. Im-proved by multi-stage feature fusion and attention techniques, the lightweight design provides a sensible answer for practical uses. Future scholars can focus on enhancing mask generation, handling various degradations, and optimizing for real-time applications, thereby strengthening the impact of the model. With great possibilities for both academic study and commercial use in computer vision, this work lays a strong basis for improving facial image restoration.